# Online Adaptive Machine Learning Based Algorithm for Implied Volatility Surface Modeling

**Yaxiong Zeng[1], Diego Klabjan[2]**

Version: June 2018

**Abstract**

In this work, we design a machine learning based method – online adaptive primal support vector regression (SVR) – to model the implied volatility surface (IVS). The algorithm proposed is the first derivation and implementation of an online primal kernel SVR. It features enhancements that allow efficient online adaptive learning by embedding the idea of local fitness and budget maintenance to dynamically update support vectors upon pattern drifts. For algorithm acceleration, we implement its most computationally intensive parts in a Field Programmable Gate Arrays hardware, where a 132x speedup over CPU is achieved during online prediction. Using intraday tick data from the E-mini S&P 500 options market, we show that the Gaussian kernel outperforms the linear kernel in regulating the size of support vectors, and that our empirical IVS algorithm beats two competing online methods with regards to model complexity and regression errors (the mean absolute percentage error of our algorithm is up to 13%). Best results are obtained at the center of the IVS grid due to its larger number of adjacent support vectors than the edges of the grid. Sensitivity analysis is also presented to demonstrate how hyper parameters affect the error rates and model complexity.

## 1. Introduction

Machine learning is gaining interest in the finance industry. In the last two decades, support vector machine, neural networks, decision trees, reinforcement learning, genetic programming and other machine learning models have been widely applied to tackle complex problems in finance, such as market direction forecasting, sentiment analysis, portfolio optimization, bankruptcy prediction, credit risk modeling, etc. For these topics, an important aspect is the challenge of the non-stationarity of noisy data, due to parameter regimes varying from time to time. Inability to react to a pattern drift can lead to damaging predictive performance and unprofitability in real-time trading. This fact motivates us to go beyond off-line training and to propose a novel online adaptive machine learning algorithm that is applied, for example, to tick data from the S&P500 options market.

---

1 PhD, Industrial Engineering and Management Sciences, Northwestern University
E-mail: yaxiongzeng2015@u.northwestern.edu
2 Professor, Industrial Engineering and Management Sciences; Director, MS in Analytics, Northwestern University
E-mail: d-klabjan@northwestern.edu

Since the inception of the Black-Scholes-Merton model, implied volatility surface (IVS) modeling has been a popular topic in options pricing theory. IVS is a mapping from the strike prices and time to maturity of options to a nonnegative value, implied volatility, whose value depends on strike prices, time to maturities, interest rates, dividends and so forth. Despite the recognition that their assumptions do not hold in a realistic trading environment, the Black-Scholes-Merton formula is widely used due to its simplification from an option price to a nonnegative value called implied volatility, which enables a fair comparison of options with different strikes, maturity and the underlying assets. As Poon and Granger (2003) point out, option implied volatility is shown to have the most information on future market volatility and outperforms classical time series based models. It also performs well across different asset classes and over a long forecasting horizon. Various methods can be used to model the IVS, such as stochastic volatility models, Levy processes, generalized autoregressive conditional heteroscedasticity models (GARCH), spline interpolation, etc. (see Homescu, 2011 for a survey). Nonetheless, machine learning algorithms are seldom applied. Recent works include Audrino and Colangelo (2010) (regression trees) and Wang, Lin et al. (2012) (artificial neural nets). Yet, all the above works view IVS modeling from a static perspective, not allowing the model to update adaptively when new market information arrives.

In this work, we propose a novel adaptive machine learning method based on support vector regression (SVR), which is further employed to update IVS. The SVR method designed is an adaptation and enhancement of Shalev-Shwartz et al. (2007), who develop an effective support vector machine (SVM) method called Primal Estimated sub-GrAdient SOlver for SVM (Pegasos) using stochastic sub-gradient descent that solely optimizes the primal objective function. Compared with the dual formulation, their primal SVM has an advantage of simplicity and can be easily adapted to the stochastic gradient descent method. Aiming at classification, they briefly mention the modification of the Pegasos algorithm suitable for regression with $\epsilon$-intensive loss but they do not derive the full regression algorithm, which we discuss in details. As an enhancement, we introduce the concept of feature vector selection (FVS) into the primal SVR algorithm. Instead of training with all data, the online algorithm updates the model using selective data points (as support vectors) that are orthogonal in the reproduced kernel Hilbert space. The idea of combining FVS and SVR is first proposed by Liu and Zio (2016), but their online SVR is based on the dual formulation of the optimization problem rather than the primal. In addition, to adaptively modify the model upon pattern drift, their solution attributes to incremental and decremental learning (Cauwenberghs and Poggio, 2001), while our algorithm updates support vectors by budget maintenance through removal (Wang, Crammer et al., 2012), which maintains the support vector size defined in FVS. To further speed up the algorithm, we implement the most computationally intensive parts in a Field Programmable Gate Arrays (FPGA) hardware developed by Maxeler Technologies, and contrast its runtime performance against a pure CPU implementation.

To summarize, our contributions focus on the following four aspects.

1. This work presents the first derivation and implementation of online primal kernel SVR. Pegasos provides an algorithm for primal kernel SVM and a quick mentioning of the extension to primal kernel SVR with no implementation and computational study in the SVR setting.
2. We provide an algorithmic enhancement to online primal SVR by means of FVS and adaptive support vector updates through budget maintenance.
3. We propose a new empirical IVS modeling algorithm using our online primal SVR.
4. A new application of the FPGA technology is provided to accelerate the most computationally intensive parts in our algorithm.

The rest of this paper is structured as follows. Section 2 reviews existing literature. Section 3 gives background of primal SVR algorithms and IVS modeling. Section 4 presents our SVR algorithm and its application to model IVS. Section 5 exhibits an empirical study using tick data from the S&P500 options market. Section 6 draws conclusions and presents future work.

## 2. Literature Review

A handful of financial applications using adaptive machine learning models have been recently developed. Chen et al. (2011) propose a bankruptcy prediction model based on an adaptive fuzzy k-nearest neighbor. The neighborhood size and the fuzzy strength parameter are updated over time by continuous particle swarm optimization. Li et al. (2012) apply an evolution strategy based support vector machine (SVM) to perform credit risk classification, which adapts the penalty term in the objective function according to time-varying data structures. Sun et al. (2013) put forward a method named adaptive and dynamic ensemble of SVM to predict corporate financial risk with focus on the concept drift of financial distress hidden in a corporate data flow. Booth (2016) explores the use of artificial neural nets, SVM, random forests and other machine learning methods in adaptive stock price return prediction and limit order book modeling. Similar to these applications, we emphasize on the ability to update the model upon occurrence of pattern drift, but our focus is on dynamic IVS modeling.

In financial market volatility forecasting, a few papers focus on SVR. Chang and Tsai (2008) introduce the combination of SVR, grey model and GARCH using artificial neural nets and show that the composite models perform better in volatility prediction than a time series method. Chen et al. (2010) apply SVR under the GARCH framework to forecast market volatility. They conclude that SVM-GARCH models are better than all competing methods in most situations of one-period-ahead forecasting. Wang (2011) combines SVR and a stochastic volatility model with jump to form an efficient currency option pricing model. He claims that the new model reduces forecasting errors and outperforms artificial neural nets. While machine learning has been widely recognized in forecasting market volatility (refer to Hahn, 2013 for a detailed survey), IVS from the Black-Scholes-Merton model has not yet been extensively studied by machine learning approaches compared to classic mathematical finance approaches, although it is gaining interests. A few examples are as follows. Malliaris and Salchenberger (1996) apply artificial neural nets to forecast S&P100 implied volatility with past volatilities and other options market factors. Fengler et al. (2007) model IVS dynamics using a semiparametric factor model by means of a principal component analysis, with empirical experiments using the DAX index options data. Lee et al. (2007) propose a particle swarm optimization method. Based on an analysis of the Korea Composite Stock Price Index (KOSPI) 200 index options market, they find that their prediction yields option prices closer to theoretical values than generic algorithms. Audrino and Colangelo (2010) present a semi-parametric model by means of regression trees to forecast implied volatility and conduct an empirical study for S&P500 index options. Wang, Lin et al. (2012) apply a neural network trained by backpropagation to forecast TXO (Taiwan Futures Exchange Option) prices under different volatility models using intraday data from 2008 to 2009. As a closely related application to IVS, recently machine learning is also increasingly being applied to option pricing (e.g. Park et al., 2014, Das and Padhy, 2017). Essentially, option prices can be derived from implied volatility but implied volatility is more general since it also implies a market panic indicator in the futures market. For this reason, option pricing does not have to rely on implied volatility predictions which is the case in the aforementioned works. These works directly regress the option prices against independent variables and circumvent the intermediate step of IVS modeling. All papers assert promising results in implied volatility or option price prediction, which further motivates us to explore an SVR application to IVS.

SVR is the regression form of SVM. Usually, SVR is formulated as a dual optimization problem. For online training of the dual, Cauwenberghs and Poggio (2001) propose incremental and decremental support vector machine that can be used to bound the number of support vectors in a model and updates the model by one support vector at a time. The increments using matrix manipulation are adiabatic, allowing the retention of Karush-Kuhn-Tucker conditions on all previous training data. In turn, the decrement step is a reversal of the increment by means of a leave-one-out procedure. Throughout the updating process, they require a book-keeping routine that migrates data points among different sets of support vectors: margin support vectors, error support vectors and (ignored) vectors within the margin. Ma et al. (2003) apply incremental and decremental SVM in a dual $\epsilon$-SVR setting (Vapnik, 1998) (named accurate online SVR). Similar to accurate online SVR, our online primal $\epsilon$-SVR algorithm entails a support vector adding and removal process, an online budget maintenance idea that is first proposed by Crammer et al. (2004) and thoroughly discussed in Wang, Crammer et al. (2012). Due to the primal setting, the model update rule requires much lower computational resources than incremental and decremental SVM. Budget maintenance of support vectors plays an important role in keeping the sparsity of an online model regardless of the primal or dual formulation; without it, the number of support vectors typically grows linearly with the number of training examples (Steinwart, 2003). In this work, we introduce the budget maintenance idea into the primal $\epsilon$-SVR algorithm called Pegasos (adapted by us from its original SVM version, proposed by Shalev-Shwartz et al., 2007). Compared with a well-established dual formulation, the primal problem is much easier and faster to solve using stochastic sub-gradient descent (Shalev-Shwartz and Srebro, 2008). Similar stochastic gradient descent based methods are applied to SVM classification problems by Kivinen et al. (2004) and Zhang (2004), who use different learning rates than Pegasos. For a detailed comparison of large scale and online SVM methods, we refer the reader to Wang, Crammer et al. (2012).

An additional challenge is how to decide the upper limit of support vectors during budget maintenance. Liu and Zio (2016) embed the idea of FVS, first proposed by Baudat and Anouar (2003), into dual $\epsilon$-SVR. Inspired by them, we include FVS in primal $\epsilon$-SVR with their notions of new pattern and changed pattern. FVS is designed for kernel implementations targeting at complexity control of the size of feature basis, in our case, the number of support vectors. To insert a new feature (or support) vector, the rule of thumb is to determine if the mapping of a new data point is nonlinearly independent from existing support vectors in the reproduced kernel Hilbert space. If so, it is viewed as a new pattern that cannot be expressed as a linear combination of the mapping of existing support vectors and is immediately added into the support vector set. Unlike a new pattern, a changed pattern indicates that the mapping of the new data point is not linearly independent in the reproduced kernel Hilbert space, but the bias of its predicted value exceeds a predetermined threshold. In this case, an existing support vector is replaced by the changed pattern while the nonlinear independence of all support vectors in reproduced kernel Hilbert space is still preserved. Continuously adding support vectors by detecting new patterns and replacing support vectors by identifying changed patterns are critical steps in our algorithm that are essential for adaptive model update, sparsity preservation and computational cost/complexity/overfitting reduction. New patterns determine the number of support vectors needed while changed patterns tell us when and where budget maintenance thought support vector removal is to be performed. A similar method that involves adaptive quantity control of support vectors is $\nu$-SVR (Schölkopf et al., 2000) that employs a different loss function than $\epsilon$-SVR. Recently, Gu et al. (2015) combine $\nu$-SVR with incremental and decremental SVM and design a new online algorithm: incremental $\nu$-SVR (INSVR). However, the decremental (support vector removal) step is missing in their work.

During the training phase of our SVR algorithm, the inverse of the kernel matrix has to be constantly updated upon support vector insertion and replacement. To accelerate such computation, we implement the matrix inverse calculation in the FPGA hardware developed by Maxeler Technologies. Besides this, the prediction part of our algorithm also has its implementation in FPGA. In existing literature, many forms of SVM have been designed specifically for a parallel FPGA implementation, with recent examples such as Coordinate Rotation Digital Computer (CORDIC) based SVM and SVR by Ruiz-Llata et al. (2010), a novel Cascade SVM by Papadonikolakis and Bouganis (2012), and an adapted Cascade SVM by Kyrkou et al. (2013). All of these papers are for inference only, due to the iterative nature of the training phase that is difficult to parallelize. We not only implement the inference in FPGA, but also parts of the training procedure.

## 3. Background

In this section, we review basics of $\epsilon$-SVR, kernel Pegasos SVR and IVS modeling.

### 3.1 $\epsilon$-SVR

Given a training set $S = \{(x_i, y_i)\}_{i=1}^m$, where $x_i \in \mathbb{R}^n, y_i \in \mathbb{R}$, $\epsilon$-SVR solves the following quadratic optimization problem

$$\min_{w,b} \frac{\lambda}{2}\|w\|^2 + \frac{1}{m}\sum_{i=1}^{m} l\big(w, b; (x_i, y_i)\big), \tag{1}$$

where the $\epsilon$ loss function is

$$l\big(w, b; (x_i, y_i)\big) = \begin{cases} |y_i - f(x_i)| - \epsilon, & |y_i - f(x_i)| \geq \epsilon, \\ 0, & \text{otherwise,} \end{cases}$$

and the estimate function

$$f(x) = \langle w, \phi(x) \rangle + b. \tag{2}$$

The L2 norm in the objective function represents the regularization term, where $\lambda$ is referred as a regularizing parameter that serves to shrink the overall model complexity. The second term is the average empirical error measured by loss function $l\big(w, b; (x_i, y_i)\big)$. Optimization in (1) penalizes data points whose $y$ values differ from $f(x)$ by more than $\epsilon$. In (2), $\phi(x)$ is a nonlinear mapping from input $x$ to reproduced kernel Hilbert space; $\langle u, v \rangle$ denotes the standard inner product between vectors $u$ and $v$; term $b$ is the regression intercept.

Estimates of $w$ and $b$ can be obtained by solving the following equivalent model to (1):

$$\min_{w,b} \frac{\lambda}{2}\|w\|^2 + \frac{1}{m}\sum_{i=1}^{m} (\xi_i + \xi_i^*)$$

subject to

$$y_i - f(x_i) \leq \epsilon + \xi_i,$$
$$f(x_i) - y_i \leq \epsilon + \xi_i^*,$$
$$\xi_i^*, \xi_i \geq 0, \qquad i = 1, \ldots, m.$$

Slack variables $\xi_i$ and $\xi_i^*$ measure the excess deviation of positive and negative errors. They are added to cope with the scenarios where no function $f(x)$ exists to satisfy the $\epsilon$ constraints by allowing regression error up to $\xi_i^*$ or $\xi_i$.

### 3.2 Kernel Pegasos SVR algorithm

The dual formulation of SVR attracted more attention than the primal, with various versions of online dual SVR proposed based on incremental and decremental SVM (refer to Section 2). In contrast, we dedicate our effort to devising a primal online SVR, enhanced from a stochastic sub-gradient descent based SVM algorithm called Pegasos (Shalev-Shwartz et al., 2007), originally for classification. Compared with the dual, the primal formulation of SVR has an advantage of simplicity and can be easily adapted to the stochastic gradient descent method. Next, we derive the regression version of the Pegasos algorithm.

The convex optimization problem (1) can be rewritten as follows by substituting the loss function into the objective:

$$\min_{w,b}[g(w,b;x_i,y_i)] = \left[\frac{\lambda}{2}\|w\|^2 + \frac{1}{m}\sum_{i=1}^{m}(\max\{0, y_i - f(x_i) - \epsilon\} + \max\{0, f(x_i) - y_i - \epsilon\})\right]. \quad (3)$$

To solve (3), the stochastic sub-gradient descent method takes one random data point $(x_i, y_i)$ at a time to estimate the sub-gradient of $g$, which reads

$$\nabla g_w = \lambda w + \begin{cases} \phi(x_i), & \text{if } f(x_i) - y_i - \epsilon > 0, \\ -\phi(x_i), & \text{if } y_i - f(x_i) - \epsilon > 0, \\ 0, & \text{otherwise.} \end{cases}$$

$$\nabla g_b = \begin{cases} 1, & \text{if } f(x_i) - y_i - \epsilon > 0, \\ -1, & \text{if } y_i - f(x_i) - \epsilon > 0, \\ 0, & \text{otherwise.} \end{cases}$$

With these sub-gradients, it is clear that the rules to update $w$ and $b$ are

$$w \leftarrow w - \frac{1}{\lambda t}\nabla g_w, \qquad b \leftarrow b - \frac{1}{\lambda t}\nabla g_b, \quad (4)$$

where $t$ represents the current iterate index, and $\frac{1}{\lambda t}$ the learning rate. Substituting $\nabla g_w$ and $\nabla g_b$ into (4), we obtain

$$w \leftarrow \left(1 - \frac{1}{t}\right) w \pm \frac{\phi(x_i)}{\lambda t}, \qquad b \leftarrow b \pm \frac{1}{\lambda t}, \quad (5)$$

if the sample falls outside the $\epsilon$ bound; otherwise,

$$w \leftarrow \left(1 - \frac{1}{t}\right) w. \quad (6)$$

One of the major benefits of SVR is the kernel trick that avoids direct access to the high-dimension mapping $\phi$ and only uses the inner products of samples specified through a kernel function. We next discuss how to embed the kernel trick with a support vector dictionary $S$. Every time a sample $x$ falls out of the $\epsilon$ bound, it becomes a support vector if it is not a current support vector; coefficient $w$ is updated by a discounted mapping $\pm\frac{\phi(x)}{\lambda t}$. This leads to creating a dictionary to keep track the cumulative sum of the discount factors $\pm\frac{1}{\lambda t}$ for each support vector. To be specific, the keys of $S$ are comprised of current support vectors and their corresponding values are the cumulative sums of the discount factors. The regression coefficient $w$ can be represented as

$$w = \sum_{s\in S} S[s] \cdot \phi(s)$$

and we have

$$f(x) = \langle w, \phi(x)\rangle + b = \sum_{s\in S} S[s] \cdot \phi(s)^T \phi(x) + b = \sum_{s\in S} S[s] \cdot K(s,x) + b, \quad (7)$$

where $K$ is a nonlinear kernel function with $K(x_i, x_j) = \phi(x_i)^T \phi(x_j)$. The kernel trick allows a feature space of arbitrary dimensionality without explicit computation of the map $\phi(x)$. As long as a function satisfies the Mercer conditions (Vapnik, 1998), it can be used as a kernel function.

The kernel Pegasos SVR (KPSVR) algorithm is exhibited in Algorithm 1. Parameter $T$ denotes the maximum number of iterations. Step 2.b uses the primal form of the estimate function (7). Step 2.c replaces $w$ by $S$ in (5) and (6). Step 2.d updates the support vector dictionary if the new sample lies outside the $\epsilon$ bound.

**Algorithm 1 – Kernel PSVR (KPSVR)**

---

1. Initialize $S = \emptyset$
2. For $t = 1, \dots, T$
    a. Input: Randomly sample $(x_t, y_t)$
    b. Output: Predict $f(x_t)$ by iterating all keys in $S$ and using the kernel trick
    $$f(x_t) \leftarrow \sum_{s \in S} S[s] \cdot K(s, x_t) + b$$
    c. $S[s] \leftarrow \left(1 - \frac{1}{t}\right) S[s]$ for all $s \in S$
    d. If $|y_t - f(x_t)| > \epsilon$, then $x_t$ is a support vector
        If key $x_t$ is in $S$, $S[x_t] \leftarrow S[x_t] \pm \frac{1}{\lambda t}$ ; else insert a key value pair, $S[x_t] \leftarrow \pm \frac{1}{\lambda t}$
        Additionally, $b \leftarrow b \pm \frac{1}{\lambda t}$

---

### 3.3 IVS modeling

The implied volatility surface (IVS) is a mapping from the strike prices κ and time to maturity $\tau$ of options to a nonnegative value – implied volatility, i.e. a mapping

$$\tilde{\sigma}_t^{IV} : (\kappa, \tau) \mapsto \mathbb{R}^+.$$

Implied volatility at a given point and measured as the standard deviation of the rate of return of the underlying asset is obtained by plugging the option price, the price of the underlying asset, the risk-free rate (estimated by Treasury yield in this paper), κ and $\tau$ into the Black-Scholes-Merton formula and back-solving for implied volatility. Since there is no closed-form solution for computing implied volatility, typical methods are by bisection or Newton-Raphson. Implied volatility is valuable for comparison of options with dissimilar characteristics such as different underlying, strike, time to maturity, etc. Although the Black-Scholes-Merton model assumes constant volatility across all options, empirical evidence shows the existence of the volatility smile and skew among a cross-section of options. Moreover, the IVS is not static; it changes over time and thus requires adaptive updates.

To model the IVS, we turn to a parametric quadratic volatility function introduced by Dumas et al. (1998). The following ad hoc model has been proven to be a simple yet robust method (usually the best among all competing functional forms) to approximate the IVS:

$$\tilde{\sigma}_t^{IV}(\kappa, \tau) = \alpha_0 + \alpha_1 \kappa + \alpha_2 \kappa^2 + \alpha_3 \tau + \alpha_4 \kappa\tau.$$

It explores the variation in volatility to asset price and time. The quadratic form is chosen due to the parabolic shape of the IVS and an attempt to avoid over-parametrization. In our kernel SVR setting, this function translates into a 4-dimension representation of each data point $(\kappa, \kappa^2, \tau, \kappa\tau)$, which can be further substituted into a kernel function to calculate the dot products between two samples.

## 4. Method

In this section, we describe FVS, budget maintenance, our enhanced kernel Pegasos SVR algorithm, its adaptation to IVS modeling and how FPGA technology is applied to accelerate the computationally intensive parts of our algorithm. The general process of our method is summarized in Figure 1.

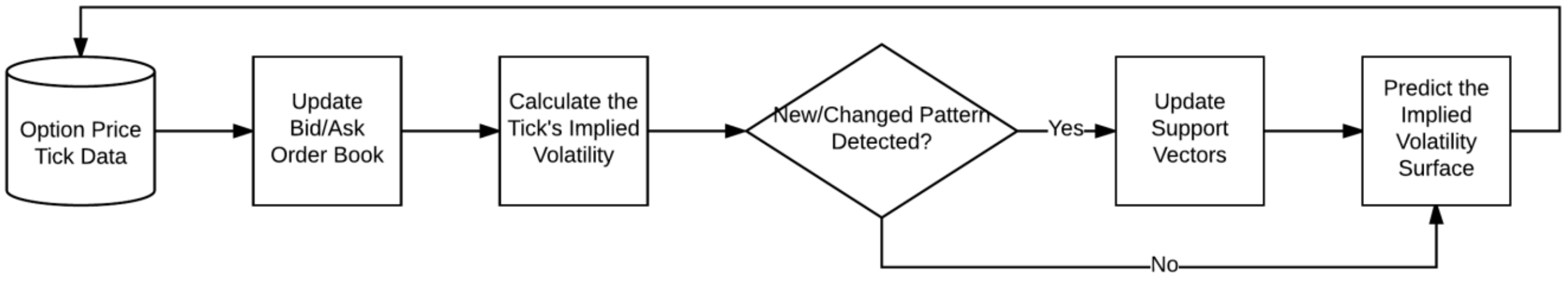

**Figure 1: Flowchart of the Method**

### 4.1 FVS

The idea of FVS is first proposed by Baudat and Anouar (2003) to select "feature" vectors from a data set and form a basis in the reproduced kernel Hilbert space that can express other data points by projection, i.e. a linear combination of the mapping of selected vectors. In our SVR setting, FVS is treated as a natural way to add support vectors and control the size of the support vector set. Furthermore, FVS is designed specifically for the kernel trick, enabling a seamless integration with SVR.

To determine if a new data point can be spanned by existing support vectors, the following statistic, named *local fitness*, is calculated:

$$J_{S,x} = \frac{k_{S,x}^T K_{S,S}^{-1} k_{S,x}}{k_{x,x}}, \tag{8}$$

where $S$ denotes the current support vector set, $x$ is a new data point, $K_{S,S}$ represents the kernel matrix, $k_{S,x}$ denotes the kernel vector of dot products between $x$ and the support vectors, $k_{x,x}$ is the dot product of $x$ mapping itself.

Local fitness functions as an approach to measure the maximum possible collinearity between the original data mapping and the approximation using a linear combination of the mapping of support vectors (Baudat and Anouar, 2003). In their original work, FVS is an iterative process of forward selection that repeatedly samples the entire data set to find the next support vector with smallest local fitness. This searching process is terminated when the maximum number of support vectors is reached, or the average local fitness of all data points (called *global fitness*) exceeds a certain threshold, or a complete basis is found. Since our SVR is an online algorithm, their framework does not fit our need. Instead we enforce a threshold criterion to enlarge the support vector set, i.e. add a new data point as a new support vector if its local fitness is smaller than a given threshold $\rho$, in which case the new data point cannot be sufficiently approximated by any linear combination of the mapping of existing support vectors (thus the invertibility of $K_{S,S}$ and its nonlinear independence from existing support vectors are guaranteed). The new data point is aliased as a *new pattern*. Note that a smaller $\rho$ leads to a lower number of support vectors, and vice versa. Choosing a good $\rho$ is hence important to help noise reduction while keeping a sufficient number of support vectors for satisfactory model performance.

**4.2 Budget maintenance**

When a new data point $x$ arrives that is not a new pattern (i.e. a large local fitness is present), we ought to further check if it represents a *changed pattern*. A changed pattern occurs if its prediction error by the current model surpasses a certain limit $\epsilon$, indicating that the support vector set needs an adaptive update: a removal of an existing support vector (old pattern) and an insertion of the new data point (changed pattern). The number of support vectors, however, remains unchanged since it is controlled by FVS.

To determine which support vector to remove, Liu and Zio (2016) put forward a contribution based method by deleting the least contributing support vector. Wang, Crammer et al. (2012) discuss three budget maintenance ideas that fix the number of support vectors to a pre-specified value $B$ (in our case, a value controlled by FVS): support vector removal, projection and merging. Support vector projection projects a support vector onto remaining support vectors while support vector merging merges two support vectors and creates a new one. In our primal setting, removal is much easier to accomplish for budget maintenance purpose, which also results in less kernel matrix manipulations than projection and merging. Since we already have the support vector dictionary $S$, we could simply remove the key with the smallest absolute value in $S$, known as a process that leads to the least gradient error or, equivalently, the least weight degradation (Wang, Crammer et al., 2012). In the following analysis, budget maintenance refers to support vector removal.

Yet this works only for the Gaussian kernel. For a general kernel function $K$, we remove the support vector key with least $S[s] \cdot \phi(s)$ or based on the kernel trick $(S[s])^2 \cdot K(s, s)$. It is easy to see that this rule in the case of Gaussian kernel, which has $K(x, x) = 1$ for any $x$, is the same as least weight degradation. The support vector dictionary we create serves two purposes: a regression coefficients container as in (7) and a reference for budget maintenance. After the old pattern has been removed, the changed pattern $x$ is added to $S$.

Every time the keys in dictionary $S$ are modified, the kernel matrix $K_{S,S}$ demands an update. More challenging, the inverse of $K_{S,S}$ in (8) needs to be recalculated. Since only one support vector is added or removed at a time in our algorithm, we must be able to efficiently manage these matrix inverse computations. Baudat and Anouar (2003) propose a method that only deals with support vector addition, which does not work directly on the inverse matrix. Nonetheless, their method cannot be extended to the case of support vector deletion because their framework continuously adds new support vectors (without deletion) when fitness requirement is not satisfied and this, in turn, allows mathematical circumvention of matrix inversion. In the following, we derive the formulas for updating the kernel inverse upon support vector addition and removal.

Suppose we have a working set of $n$ support vectors forming kernel matrix $K_n$ (we leave out subscript $S,S$ for notation simplicity) and its inverse $K_n^{-1}$. Let us assume we want to add a new support vector $x$, and update the kernel matrix by appending a new column and a new row:

$$K_{n+1} = \begin{pmatrix} K_n & k_{S,x} \\ k_{S,x}^T & k_{x,x} \end{pmatrix},$$

where $k_{S,x}$ denotes an $n \times 1$ vector of dot products between $x$ and the previous $n$ support vectors.

Let

$$K_{n+1}^{-1} = \begin{pmatrix} X & Y \\ Y^T & z \end{pmatrix} \tag{9}$$

be the updated inverse matrix, where $X$ is an $n \times n$ matrix, $Y$ is an $n \times 1$ vector, and $z$ is a scalar. The inverse matrix is symmetric because the kernel matrix is always symmetric.

The solutions for $X, Y$ and $z$ are

$$\begin{aligned} z &= 1/\left(k_{x,x} - k_{S,x}^T K_n^{-1} k_{S,x}\right), \\ Y &= -z K_n^{-1} k_{S,x}, \\ X &= K_n^{-1} - K_n^{-1} k_{S,x} Y^T. \end{aligned} \tag{10}$$

The denominator of $z$ is never zero, because otherwise, the local fitness of the new data point $x$ is one, meaning it is an existing support vector and cannot be inserted into the support vector dictionary once again (assuming $\rho < 1$, which is reasonable).

Upon support vector deletion, we are given $K_{n+1}^{-1}$ as in (9) and after deleting $Y$ we obtain:

$$K_n^{-1} = X - \frac{YY^T}{z}. \tag{11}$$

Updating the kernel matrix is straightforward by removing the row and column corresponding to the deleted support vector.

### 4.3 Enhanced KPSVR algorithm

We first incorporate budget maintenance and FVS to KPSVR. Wang, Crammer et al. (2012) discuss how budgeted SVM can improve computational efficiency in both time and space, but with no mentioning of its potential extension to SVR. Algorithm 2 exhibits the budgeted KPSVR algorithm. Step 2.e removes the support vector with the least absolute value if the maximum number is exceeded. Budget parameter $B$ should be carefully chosen based on trade-off between prediction accuracy and practical limitations such as memory, speed and data throughput.

**Algorithm 2 – Budgeted KPSVR (BKPSVR)**

1. Initialize $S = \emptyset$
2. For $t = 1, \ldots, T$

   Step 2.a to 2.d from KPSVR

   e. If $|S| > B$, select the key $s$ in $S$ with the smallest $(S[s])^2 \cdot K(s, s)$, remove its key value pair

A fixed number of support vectors may not be optimal when a new pattern emerges or when data patterns are continuously changing, in which cases the number of support vectors should adapt responsively. This is addressed by incorporating FVS into BKPSVR. Once a sample $x$ cannot be sufficiently approximated by any linear combinations of the mapping of existing support vectors (i.e. a small local fitness $J_{S,x}$ that is less than the preset threshold $\rho$), it is added into the support vector dictionary $S$ as a new pattern without checking if it is a changed pattern. Otherwise, if its prediction is not within the $\epsilon$ bound, we call it a changed pattern that further activates budget maintenance. Algorithm 3 presents the enhanced KPSVR algorithm with FVS and adaptive updates of support vectors through budget maintenance. In particular, Step 2.d is modified to detect new patterns and changed patterns, where support vector addition and budget maintenance are conducted. Upon support vector insertion into or deletion from $S$, formulas (9) and (11) are used to efficiently update the matrix inverse $K_{S,S}^{-1}$ so that it is ready to calculate local fitness $J_{S,x}$ in the next iteration.

**Algorithm 3 – Enhanced KPSVR (EKPSVR)**

1. Initialize $S = \emptyset$
2. For $t = 1, \ldots, T$

   Step 2.a to 2.c from KPSVR

   d. If local fitness is violated, i.e. $J_{S,x_t} < \rho$, then $x_t$ is a new support vector (new pattern)

      Add key $x_t$ into $S$, $S[x_t] \leftarrow \pm \frac{1}{\lambda t}$, $b \leftarrow b \pm \frac{1}{\lambda t}$

   Else if $|y_t - f(x_t)| > \epsilon$, then $x_t$ is a support vector (changed pattern)

      If key $x_t$ is in $S$, $S[x_t] \leftarrow S[x_t] \pm \frac{1}{\lambda t}$ ; else select the key $s$ in $S$ with the smallest $(S[s])^2 \cdot K(s, s)$, remove its key value pair, then insert key $x_t$, $S[x_t] \leftarrow \pm \frac{1}{\lambda t}$

      Additionally, $b \leftarrow b \pm \frac{1}{\lambda t}$

### 4.4 IVS modeling by EKPSVR

To model the constantly fluctuating IVS, the stochastic gradient descent based EKPSVR algorithm has to be tailored to reflect the online nature of the training and predicting process using market data.

If EKPSVR is directly applied to model IVS without further modification, then later in time when $t$ gets large enough, newly received data barely influences the model (with small step size $\pm\frac{1}{\lambda t}$ close to 0), which is then almost unchanged and fails to capture regime changes. As a result, reopening is necessary by reinitiating $t = 1$ at the end of a certain interval for adjustments to latest market conditions. For instance, the interval might be market opening or based on empirical evidence that uses minute level frequency in the context of intraday tick data. We name such an interval a *reopening interval*. To inherit models from previous intervals upon reopening, we adjust the learning rate from $\frac{1}{\lambda t}$ to $\frac{1}{\lambda(t+\omega)}$ by introducing a positive warm-start hyper-parameter $\omega$ into the denominator (otherwise the model would be completely retrained since Step 2.c would have $1 - \frac{1}{t} = 0$).

Algorithm 4 finalizes the online IV-EKPSVR algorithm. Upon arrival of a new tick, regression errors are recorded and the model is updated according to local fitness and prediction bias, after which a new IVS prediction is made. Once reaching the end of a reopening interval, $t$ is reset to 1.

**Algorithm 4 – Online IVS-EKPSVR**

1. Initialize $t = 1, S = \emptyset$
2. Loop
    a. Input: a new observation $(x_t, y_t)$ at time $t$ (where $x_t$ is the feature vector and $y_t$ is the computed IV value based on $x_t$)
    b. Obtain $f(x_t)$ from the predicted IVS
    c. $S[s] \leftarrow \left(1 - \frac{1}{t+\omega}\right) S[s]$ for all $s \in S$
    d. If local fitness is violated, i.e. $J_{S,x_t} < \rho$, then $x_t$ is a new support vector (new pattern)
        Add key $x_t$ into $S$, $S[x_t] \leftarrow \pm\frac{1}{\lambda(t+\omega)}$, $b \leftarrow b \pm \frac{1}{\lambda(t+\omega)}$
       Else if $|y_t - f(x_t)| > \epsilon$, then $x_t$ is a support vector (changed pattern)
        If key $x_t$ is in $S$, $S[x_t] \leftarrow S[x_t] \pm \frac{1}{\lambda(t+\omega)}$; else select the key $s$ in $S$ with the smallest $(S[s])^2 \cdot K(s,s)$, remove its key value pair, then insert key $x_t$, $S[x_t] \leftarrow \pm\frac{1}{\lambda(t+\omega)}$
        Additionally, $b \leftarrow b \pm \frac{1}{\lambda(t+\omega)}$
    e. Output: IVS for each strike price and maturity of interest by (7) with the updated support vector dictionary (prediction/inference)
    f. $t \leftarrow t + 1$
    g. If the end of current reopening interval is reached, reset $t = 1$

### 4.5 FPGA implementation

Field Programmable Gate Arrays (FPGA) hardware has been widely used in the high frequency trading sector to accelerate and reduce the latency of packet capture (e.g. FIX/FAST messages), order book modeling, theoretical price calculations and other finance statistic evaluations (e.g. option greeks). In this work, an FPGA embedded server MaxWorkstation10G (developed by Maxeler Technologies) is adopted for parallel computing. This powerful server is equipped with one Vectis dataflow engine (DFE) and Intel Core i7 quad-core CPU with 16GB RAM. The Vectis board includes a Xilinx Virtex-6 SX475T FPGA, where highly parallelizable computations are performed. MaxWorkstation10G is a connectivity development platform with CPU and DFE connected via PCI Express gen2 x8, guaranteeing its ultra-low latency (Figure 2). Compared with Graphical Processing Unit (GPU) based accelerators, reconfigurable FPGA implementations enjoy lower power consumption alongside its strong capability in high-performance computing, but at the expense of ease of programming. Unlike conventional FPGAs, Maxeler FPGA solutions offer extended flexibility and significantly improve the programming experience (with MaxIDE in MaxelerOS). They developed a customized Java-based language to program the DFE kernels, which specifies computational logic, and the DFE managers, which connect the data flows among CPU, kernels, and memories (fast on-chip memory FMem or large off-chip memory LMem). Note that data flows are streamed into and out of the DFE, meaning that additional data handling is necessary, for example, matrix serialization. By Simple Live CPU interface (SLiC), the FPGA application can be embedded into a number of major programming languages such as C/C++, Java, python, MATLAB, R etc.

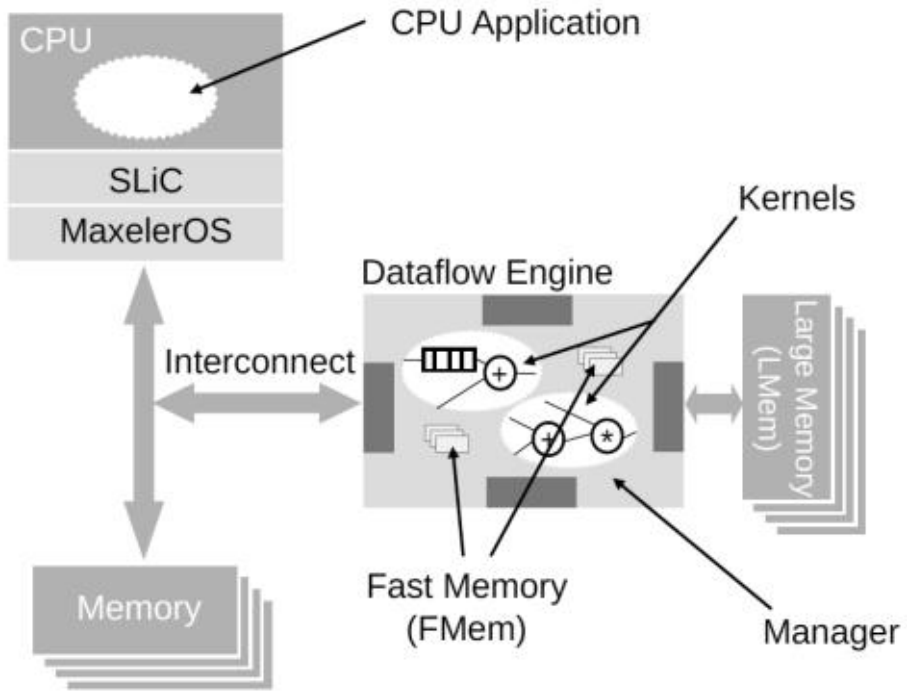

**Figure 2: Maxeler Dataflow Engine Architecture (Courtesy of Maxeler Tutorial)**

As presented in Section 4.2, the support vector insertion and deletion in Step 2.d of IVS-EKPSVR require kernel matrix inverse updates by formulas (9) and (11), which are computationally intensive and thus become a good candidate for FPGA acceleration. Essentially, these matrix updates are a number of nested-for loops that are highly parallelizable. We also notice that the local fitness calculation (at the beginning of Step 2.d) and the predictions for each sample (Step 2.e) consist only of nested-for loops. Based on these observations, four parts of the IVS-EKPSVR algorithm are prime candidates for DFE: support vector addition, support vector removal, local fitness, and sample prediction.

Consider inference (Step 2.e). To make these predictions, two for loops are required – one sweeps all data points and another scans all support vectors. Its pseudo code is as follows in Algorithm 5 (denote the prediction output vector as $p$, the number of samples as $M$, the number of support vectors as $N$).

**Algorithm 5 – DFE implementation of inference**

1. Initialize $p[i] = 0$ for all $i = 1, \dots, M$
2. Input: $S$, $K_{S,S}$
3. For $i$ from 1 to $M$
   For $j$ from 1 to $N$
   $p[i] = p[i] + S[j] \cdot K_{S,S}[i,j]$
4. Output: $p$

Our DFE implementation is shown in Figure 3 (assuming there are 3 support vectors). The outer loop in the CPU code is the target of parallelization since the predictions for each sample are independent. The support vector related data is stored in the read-only memory (ROM) in FPGA. Each time a new sample is streamed into DFE, it is then distributed into every ROM and multiplied with the support vector to calculate $S[j] \cdot K_{S,S}[i,j]$. Then, we sum all these contributions to obtain the prediction output for this particular data point.

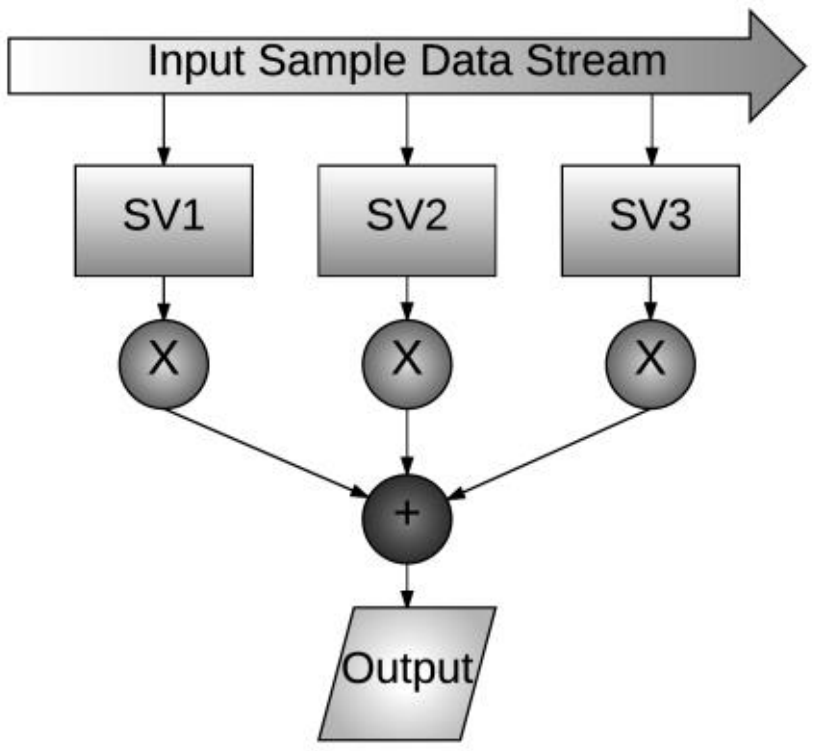

**Figure 3: DFE Design for Sample Prediction (for 3 support vectors)**

Next we discuss local fitness. The formulas to calculate $J_{S,x}$ in (8) and $z$ in (10) are very similar. In (8), the numerator is matrix multiplication $k_{S,x}^T K_{S,S}^{-1} k_{S,x}$ and the denominator is a scalar. In (10), the solution to $z$ also involves a matrix multiplication, similar to $k_{S,x}^T K_{S,S}^{-1} k_{S,x}$. They can indeed share the same DFE design. We develop a two-step data flow for this matrix multiplication: multiply $k_{S,x}^T K_{S,S}^{-1}$ first, resulting in an intermediate row vector denoted $I$; then multiply $I \cdot k_{S,x}$ and output the desired scalar $c$. The pseudo code is given below in Algorithm 6.

**Algorithm 6 – DFE implementation of local fitness and $z$ calculation**

1. Initialize $I[i] = 0$ for all $i = 1, \dots, N$
2. Initialize $c = 0$
3. Input: $k_{S,x}$, $K_{S,S}$
4. For $i$ from 1 to $N$

   For $j$ from 1 to $N$

   $I[i] = I[i] + k_{S,x}[j] \cdot K_{S,S}^{-1}[i,j]$

   $c = c + I[i] \cdot k_{S,x}[i]$
5. Output: $c$

The corresponding DFE implementation is presented in Figure 4. In both the left and right DFE kernels, we map the vector $k_{S,x}$ into on-chip ROM, meaning each box in the figure represents an element of $k_{S,x}$ (suppose the kernel inverse matrix is 3-by-3). The inverse matrix columns of $K_{S,S}^{-1}$ are then streamed into the left DFE kernel and multiplied with each element of $k_{S,x}$, which yields $k_{S,x}[j] \cdot K_{S,S}^{-1}[i,j]$. By summing up these multiplications, the intermediate row vector $I$ is attained. Vector $I$ is then streamed into the next DFE kernel on the right and multiplied with each vector element of $k_{S,x}$ to obtain $I[i] \cdot k_{S,x}[i]$. Finally, summing these multiplications yields the desired scalar $c$.

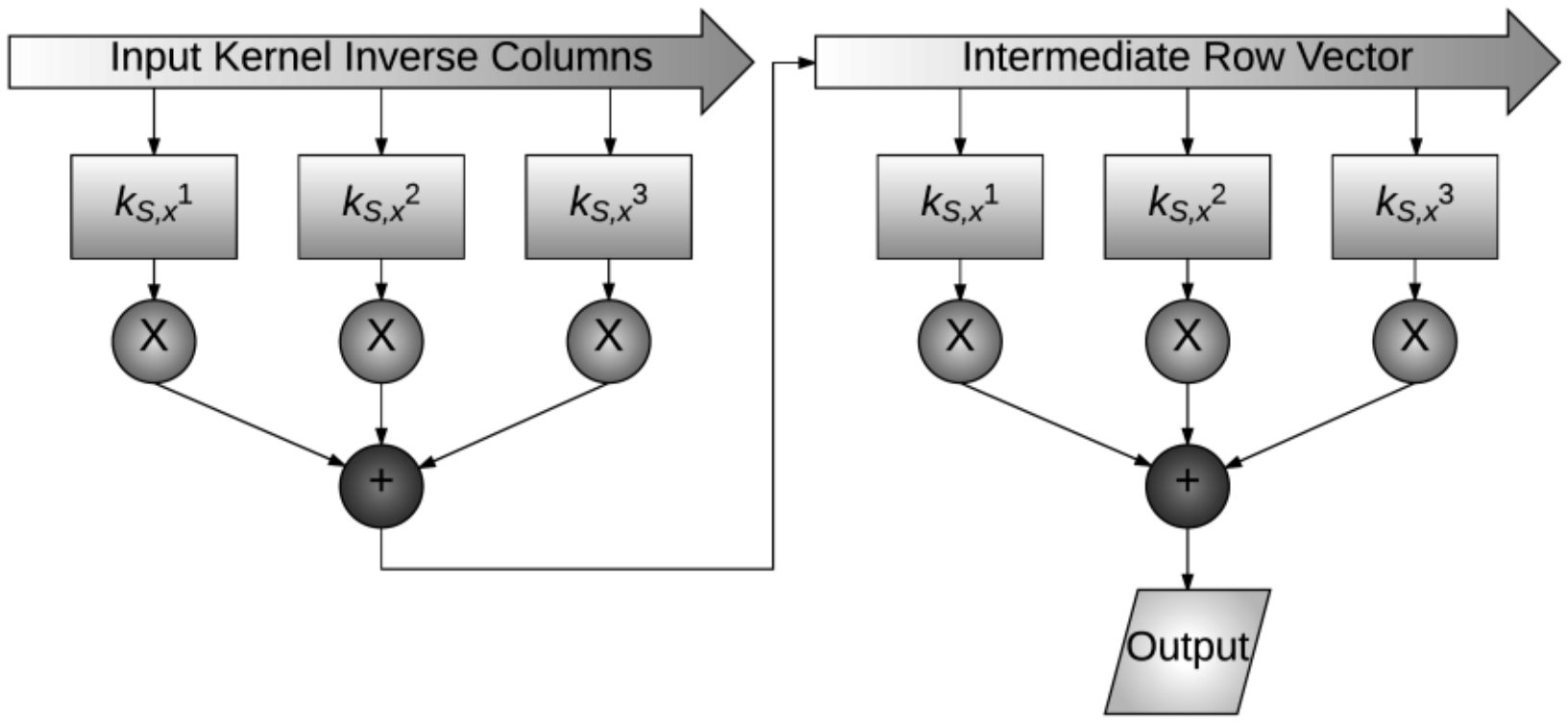

**Figure 4: DFE Design for Two-step Matrix Multiplication (for 3 support vectors)**

Once the intermediate row vector $I$ and scalar $z$ are obtained, vector $Y$ in (9) immediately follows by multiplying $I^T$ with $-z$.

Finally we discuss support vector deletion and addition. This amounts to computing $X$ in (10) and the inverse matrix update upon support vector deletion using (11). Note that the solution to $X$ in (10) requires the intermediate row vector as well, therefore we rewrite

$$X = K_n^{-1} - I^T Y^T,$$

a form identical to (11). This means that these two parts can share the same data flow design again. The pseudo code and DFE implementation are omitted due to its resemblance to the previous two cases. The basic idea is that for each element in matrix $X$ or $K_n^{-1}$, depending on the formulas, we subtract its value by the product of corresponding elements in vectors $I^T$ and $Y^T$ or the product of scalar $1/z$ and elements in vectors $Y$ and $Y^T$. Matrix $X$ or $K_n^{-1}$ is streamed into the DFE, while two vectors and a scalar (in the $X$ case, the scalar is 1) are mapped onto the on-chip ROM.

## 5. Computational Study

Using empirical data from the E-mini S&P 500 futures and options market, in this section we present a computational study that compares our IVS-EKPSVR algorithm against competing methods.

### 5.1 Data

The E-mini S&P 500 option has the E-mini S&P 500 future as the underlying asset, both traded in the Chicago Mercantile Exchange (CME). The options tick data used for this study is based on dates 01/27/2014 to 01/31/2014 and contains 5 maturities: February to June 2014. Each tick represents the latest top level of a limit order book. The trading hours are Sunday to Friday 5 pm to 4 pm Central Time with a halt from 3:15 pm to 3:30 pm, and a 60-minute break beginning at 4 pm. These non-trading time periods are excluded from our experiments. Although trading activity can occur almost any time in a trading day, the busiest hours are from 9 am to 4 pm. For example, over 54.4 million ticks are recorded during this time period out of 79.9 million on 01/27/2014, i.e. approximately 70% ticks in 30% time of a day. For this reason we built our models only for these hours. The moneyness of options is defined as the ratio of strike price divided by the underlying asset price. Because out-the-money and in-the-money options are less traded in a high-frequency intra-day setup, we limit the moneyness of options to 0.95 to 1.05 (i.e. at-the-money or ATM). Their matching strike prices are determined by the settlement price of the underlying futures in the previous trading day of 01/27/2014, which is 01/24/2014. The total number of data points on each modeled IVS (Call Bid, Call Ask, Put Bid and Put Ask) is 200, i.e. 40 strike prices of the ATM options for each of the 5 maturities. These samples on the strike-maturity grid display varying values of implied volatility over time and thus become the targets of our online prediction models.

Given the price data, implied volatilities are computed using the Black-Scholes formula with interest rates linearly interpolated from the daily Treasury yield curve. Figure 5 summarizes the statistics of the average implied volatility from 9 am to 4 pm on 01/27/2014 (details can be found in Appendix). Generally, Feb 2014 maturity shows the most volatile properties with the largest standard deviations. The longer the maturity, the smaller the standard deviation. Another observation is that put options are on average priced higher than call options by examining the mean of implied volatility.

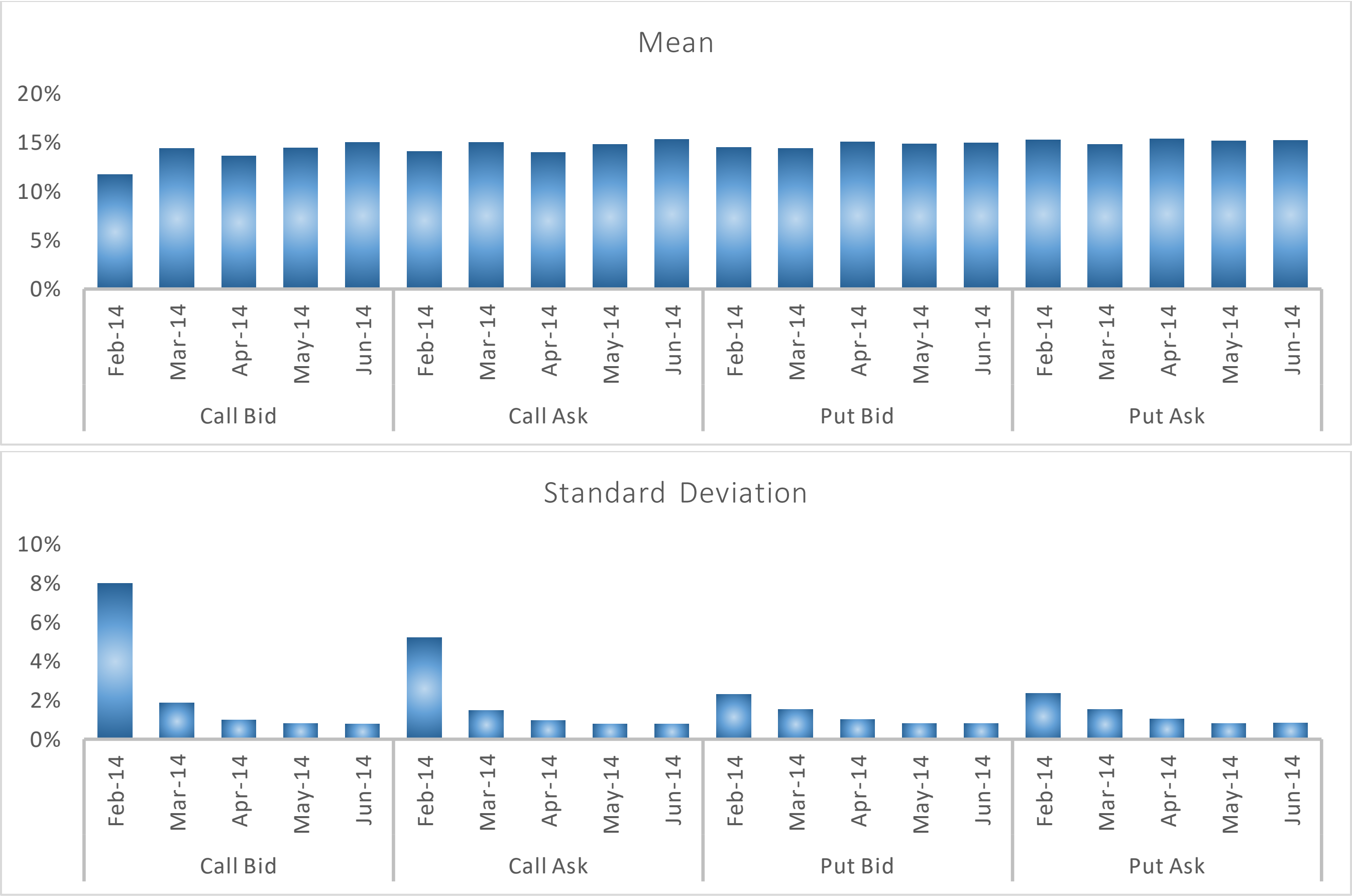

**Figure 5: Summary Statistics of IV**

Figure 6 presents the Call Bid, Call Ask, Put Bid and Put Ask IVS models for ATM options at 9 am on 01/27/2014 (strike prices range from 1,670 to 1,865 with a discrete increment of 5). Volatility smile can hardly be identified but volatility skew exists in all four surfaces. It can also be observed that dramatic changes of implied volatility appear on the higher end of strike prices for shorter maturities. The above observations not only apply to 01/27/2014, but all other four days.

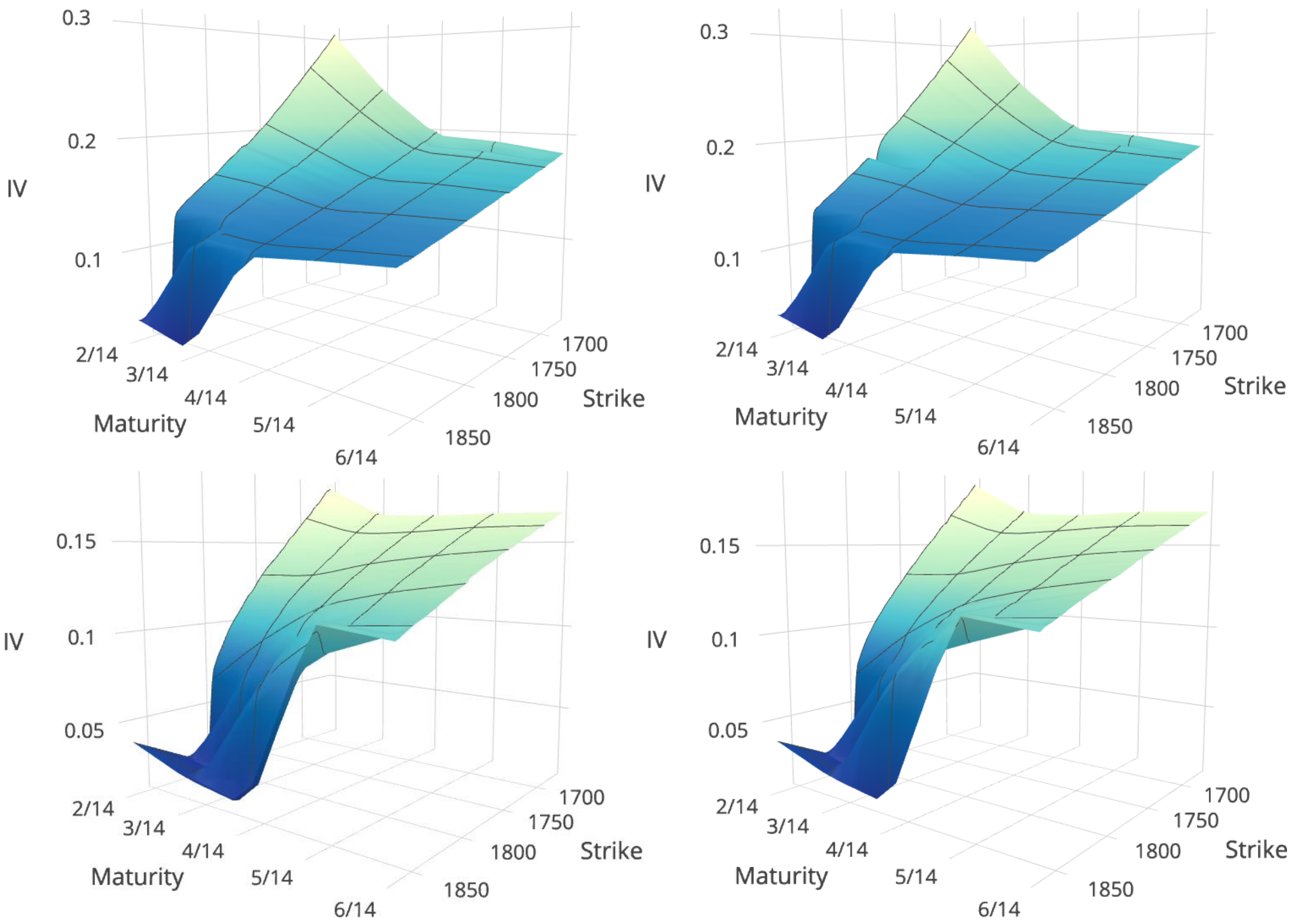

**Figure 6: IVS** (From left to right, top to bottom: Call Bid, Call Ask, Put Bid, Put Ask)

**5.2 Results**

Our algorithms are developed in C++ and a Java-based FPGA programming language on a load-free MaxWorkstation10G with 4.0 GHz Intel Core i7 processor and 16 GB of RAM.

In the implementation, once the top level of a limit order book is updated by a new tick in C++ (on the CPU), that tick is then streamed into the FPGA hardware (or DFE) to update the SVR model by examining its local fitness, updating the kernel matrix inverse upon support vector addition or removal and finally to predict the entire IVS (see Section 4.5 for details). In between these operations, C++ functions as a data transfer medium that serializes, recovers and stores vectors and matrices, and connects input and output data flows with the PCI Express portal of the FPGA hardware. A counterpart implementation that accomplishes the same computations exclusively in C++ has also been developed for a comparison purpose (we call it a pure CPU implementation).

Experiments are implemented on a slightly modified version of the online IVS-EKPSVR. In Algorithm 4, models are continuously updated at each tick and so does the IVS prediction. Empirically, tick-by-tick prediction and evaluation are not necessary since the surface does not vary drastically within a short period. This fact motivates us to delay prediction and error evaluation until receipt of a certain number of ticks or elapse of a certain amount of time. For simplicity, we set such time point to be the end of a reopening interval, where regression errors are recorded and prediction for the next interval is performed. Due to our intraday setting, the length of reopening intervals is set to be 1 minute, a lower limit of common choices between 1 to 5 minutes (Hansen and Lunde, 2005) due to higher liquidity of ATM options. Similar minute-level intervals are used by Bollerslev et al. (2009) and Sévi (2014) for estimation of volatilities using tick data. Since the IVS prediction is delayed, Step 2.b of Algorithm 4 then follows Step 2.b of KPSVR to obtain the predicted value of a new observation using the latest support vector dictionary instead of extracting it from previously predicted IVS.The hyper-parameters chosen for the IVS-EKPSVR algorithm are as follows: $\rho = 0.3, \lambda = 0.75, \omega = 7, \epsilon = 0.01$. SVR is equipped with the Gaussian kernel[3] using $\gamma = 0.25$. They were selected based on calibration with data from 01/27/2014. After fixing these hyper-parameters, the other four days essentially form a hold-out set used to assess all algorithms. We do not carry a model from the previous day to the next, i.e. the training process is restarted every morning, but the order book is constantly updated upon receipt of new ticks. Model fitting starts at 8 am, and inference begins at 9 am. This one-hour lag provides a warm start for inference tasks each day. Both model update and prediction end at 4 pm daily with a halt from 3:15 to 3:30 pm.

To better illustrate the use case of IVS-EKPSVR, we provide the following numerical example. Suppose the current time is 10:00 am and the Call Bid model has 100 support vectors (same logic applies to Call Ask, Put Bid and Put Ask models). A new sample arrives with strike price 100 and time to maturity 0.1 years, i.e. feature vector $x_t = (100, 100^2, 0.1, 100 \cdot 0.1)$ and its implied volatility $y_t = 0.2$ (Step 2.a in Algorithm 4). Since this is an online algorithm, every time a new sample is received, the values in the support vector dictionary $S$ need to be discounted based on the current time step $t$ . Suppose $t = 20$ and given $\omega = 7$, the values in the dictionary are then multiplied by $\left(1 - \frac{1}{t+\omega}\right) = \frac{26}{27}$ (Step 2.c). The next step is to determine whether this new sample is a non-existing support vector (Step 2.d). Let us assume that we find its local fitness is $0.5 > \rho = 0.3$, i.e. it is not a new pattern. Then we check if it is a changed pattern by comparing the difference of the predicted $f(x_t)$ (from Step 2.b) and the actual value $y_t$. Let us assume that prediction $f(x_t) = 0.25$, and thus $|y_t - f(x_t)| = 0.05 > \epsilon = 0.01$. We can now conclude that this sample is a changed pattern and should be inserted into the support vector dictionary $S$. If it already exists in the dictionary, value $\frac{1}{\lambda(t+\omega)} = \frac{1}{0.75(20+7)}$ should then be added to its existing value. Otherwise, we look for key $s$ in $S$ with smallest $(S[s])^2 \cdot K(s, s)$ and replace $s$ with the new support vector and its corresponding value $\frac{1}{0.75(20+7)}$. Lastly, the regression intercept term $b$ is incremented by $\frac{1}{0.75(20+7)}$. Once the interval is reopened at 10:01 am, $t$ is reset to 0.

As a simplification, KPSVR provides a baseline that excludes any enhancements, and BKPSVR embeds budget maintenance. Similar to Algorithm 4, identical IVS prediction schemas are attached to these benchmarks, naming them to IVS-KPSVR and IVS-BKPSVR correspondingly. Using the previous illustrative example, when the same sample arrives, IVS-KPSVR does not check if the local fitness condition is violated, but directly compares the prediction and actual value. Recall that this sample is a changed pattern and let us suppose it does not yet exist in the support vector dictionary. In this case, IVS-KPSVR does not look for an existing support vector to remove but directly adds the sample into the dictionary. Hence, under IVS-KPSVR the support vector size increases rather fast. On the other hand, IVS-BKPSVR sets an upper bound on the support vector size. When the sample arrives, the only difference from IVS-KPSVR is that it checks if the support vector size reaches the upper bound $B$. If yes, the algorithm removes the key with smallest $(S[s])^2 \cdot K(s,s)$ and keeps the support vector size unchanged. A careful selection of value $B$ is thus critical under IVS-BKPSVR.

[3] Gaussian kernel function: $K(x, y) = exp(-\gamma|x - y|^2)$.

We first present select behavior on the validation date of 01/27/2014. Figure 7 exhibits the implied volatility prediction time series of IVS-EKPSVR for options with a strike price of 1,770 (right ATM) on this date. The sequence-axis represents the time sequence discretized by 1 minute from 9 am to 4 pm (3:15 to 3:30 pm excluded) with 0 representing 9 am. Models from our algorithm adaptively adjust themselves and lead to predictions varying with the shifting market conditions. Specifically, prediction errors behave as if they are white noises (corresponding plots are omitted here), with absolute prediction error averaging 0.87%, 0.86%, 0.68%, 0.68% for Call Bid, Call Ask, Put Bid and Put Ask IVS models, respectively. Prediction error is relatively large on the edge of the strike-maturity grid for options with the shortest and longest maturities, because these options have less reference points to infer their implied volatilities during online learning. The aforementioned observations not only apply to strike 1,770, but also to other strikes and days.

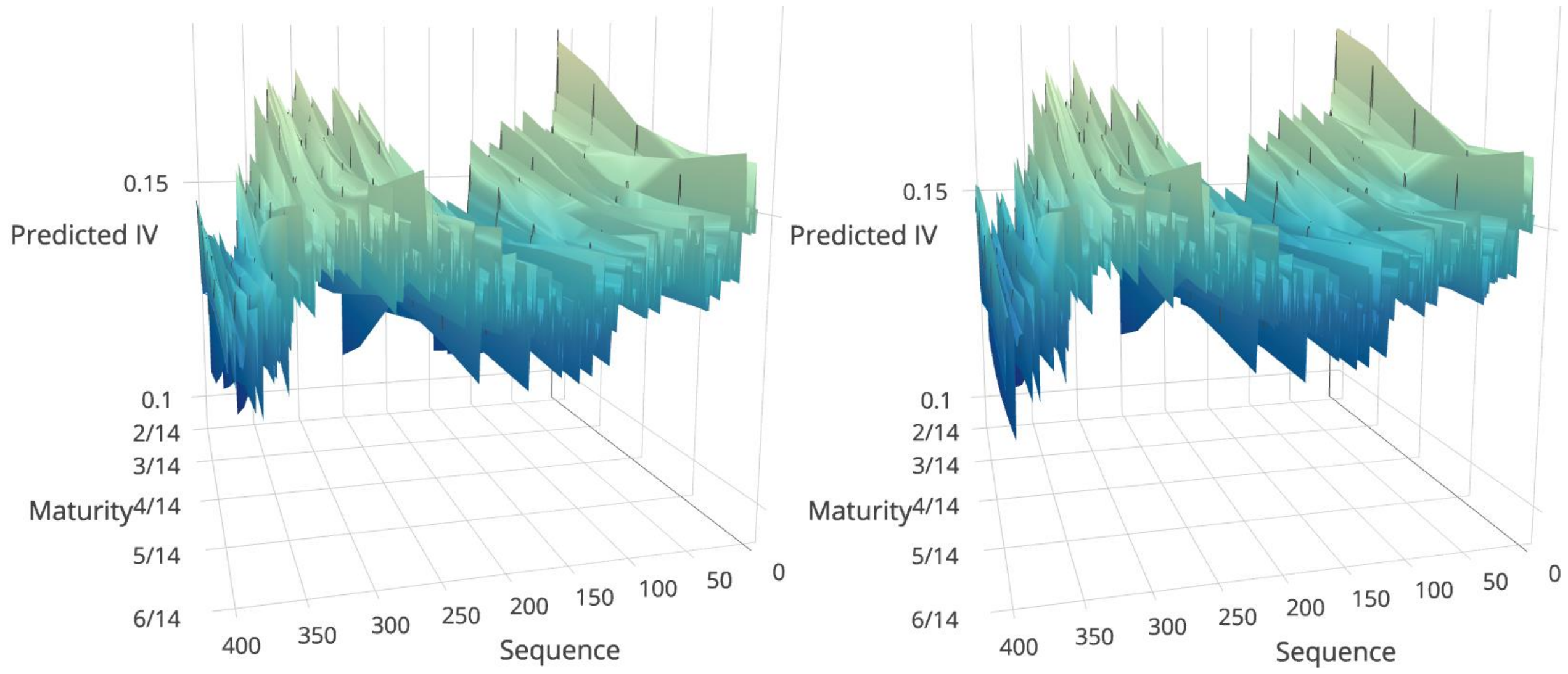

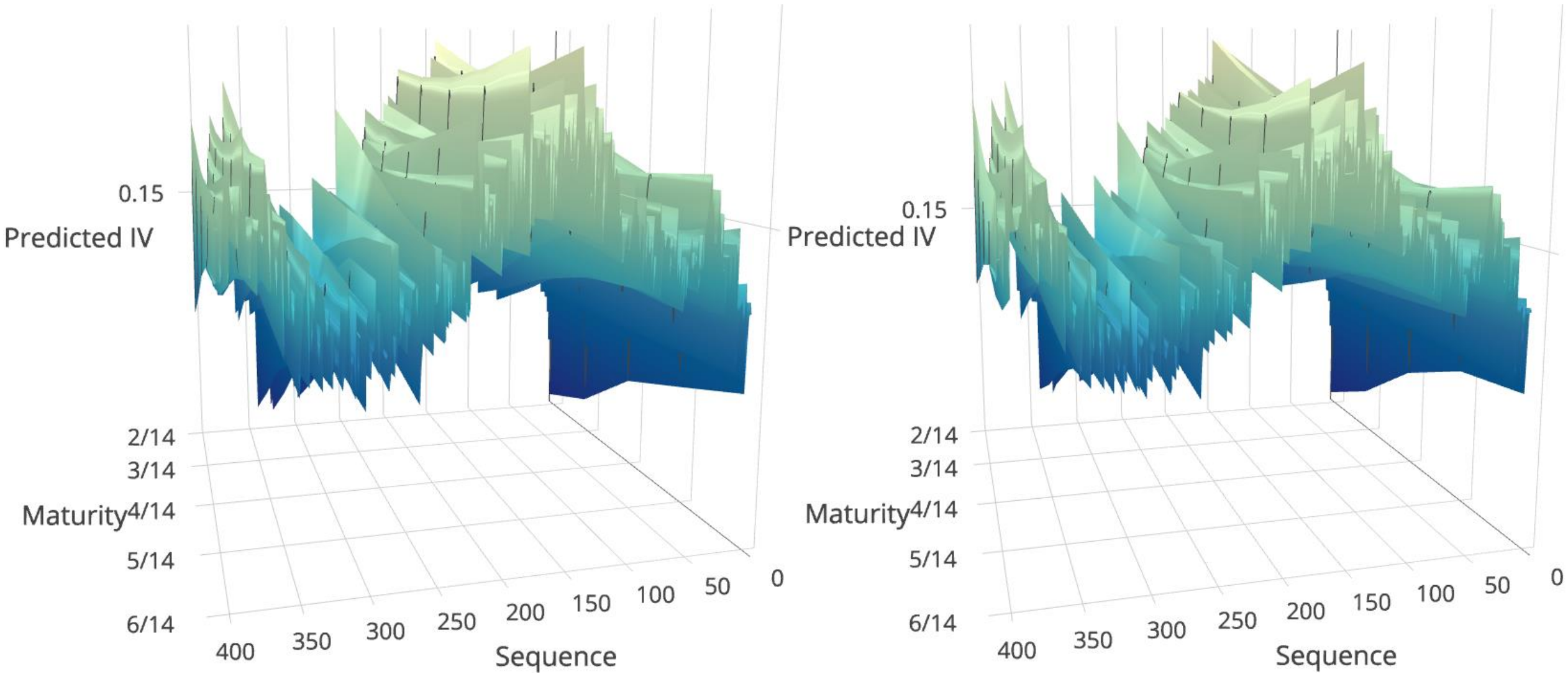

**Figure 7: IV Prediction** (From left to right, top to bottom: Call Bid, Call Ask, Put Bid, Put Ask)

We now explore a few competing models and contrast their average performance against IVS-EKPSVR with data from 01/28/2014 to 01/31/2014. For consistency, the same strike-maturity grid is used for these four days as on 01/27/2014. Average performance statistics across the four days are shown in Table 1 and Figure 8 (details can be found in the Appendix). Table 1 is about the average support vector sizes over the four days. Note that for budget maintenance purpose, the support vector size in IVS-BKPSVR is set to 50. We choose this value as a complement to the other two algorithms to reveal the performance when a quarter of available data points are considered support vectors while the other two discuss scenarios where all or half samples are used as support vectors. Figure 8 presents the differences of performance measured by average minute-by-minute MAPE (mean absolute percentage error) and RMSE (root mean square error) over the four days between IVS-EKPSVR and IVS-KPSVR or IVS-BKPSVR.

Figure 8 particularly asserts that IVS-EKPSVR and IVS-KPSVR are comparable in performance metrics with the latter slightly outperforming the former. Both clearly substantially beat IVS-BKPSVR. To verify that IVS-EKPSVR and IVS-KPSVR perform equivalently in a statistical sense, we perform a two-sample $t$ test with the null hypothesis that these two algorithms have the same mean MAPE. Using the Gaussian kernel, the $p$ values are 82.51%, 49.89%, 52.90% and 74.79% for Call Bid, Call Ask, Put Bid and Put Ask IVS models respectively, meaning that the null hypothesis cannot be rejected. The same test between IVS-EKPSVR and IVS-BKPSVR yields $p$ values of 0.91%, 0.09%, 3.37% and 1.86% implying that the mean of IVS-EKPSVR is lower than the mean of IVS-BKPSVR with 95% confidence. Analogous conclusions can be drawn for the linear kernel. However, peeking at Table 1 we observe that IVS-EKPSVR uses substantially fewer support vectors in the Gaussian kernel case than IVS-KPSVR, which does not bound the total number of support vectors and keeps adding support vectors provided that the $\epsilon$ condition is violated. Thus, in the call bid and ask models, the number of support vectors reaches the maximum 200 (the total number of data points on the strike-maturity grid, i.e. all samples) while the put bid and ask models arrive at 199. FVS in IVS-EKPSVR functions as a support vector size controller. It only adds a support vector when local fitness $J_{S,x}$ of a new sample $x$ is smaller than the preset threshold $\rho$. In the Gaussian kernel case, IVS-EKPSVR uses around 50 ~ 60% of support vectors as in IVS-KPSVR, but results in a similar performance that can be seen from Figure 8. Nevertheless, it distinguishes itself from IVS-BKPSVR by dynamic support vector size tuning and further reduction in error rates. In essence, it finds a balance point where it stops increasing model complexity once performance reaches its limit. In the linear kernel case the difference is not that pronounced. We now conclude that IVS-EKPSVR achieves the same performance numbers as the best of the two competing algorithms but with fewer support vectors and is thus the preferred choice of the algorithm.

The actual MAPE values in all settings range from 12% to 15% while RMSE is from 1.5% to 2.5%.

Taking a closer look at Table 1 and Figure 8, we obtain the following observation that under IVS-EKPSVR, the Gaussian kernel behaves better than the linear kernel, whereas the linear kernel acts analogous to the baseline IVS-KPSVR, encompassing almost entire samples into the support vector space and achieving similar performance to IVS-KPSVR. Table 2 exhibits the two-sample $t$ statistics and $p$ values of MAPE, with a null hypothesis that the linear and Gaussian kernel share the same mean MAPE. By examining Table 2, we observe that when support vector size is the same, the linear kernel performs comparable to the Gaussian kernel in IVS-KPSVR and IVS-BKPSVR since a large $p$ value does not reject the null hypothesis. With much more support vectors, the linear kernel does not lead to significant improvement over the Gaussian kernel in IVS-EKPSVR. This being said, the Gaussian kernel works better for our online algorithm.

As a final note based on Table 1 and Figure 8, we find that a larger number of support vectors leads to smaller MAPE and RMSE but at the expense of more computational resources to perform kernel matrix manipulations, a conclusion that can be further justified by the sensitivity analysis presented in the latter part of this section.

**Table 1: Support Vector Size**

| | Kernel | Call Bid | Call Ask | Put Bid | Put Ask |
|---|---|---|---|---|---|
| IVS-KPSVR | Gaussian | 200 | 200 | 199 | 199 |
| | Linear | 200 | 200 | 199 | 199 |
| IVS-BKPSVR | Gaussian | 50 | 50 | 50 | 50 |
| | Linear | 50 | 50 | 50 | 50 |
| IVS-EKPSVR | Gaussian | 110 | 104 | 116 | 120 |
| | Linear | 191 | 194 | 196 | 196 |

**Table 2: Two-sample t-test of MAPE for Gaussian vs. Linear Kernels**

| | Call Bid | | Call Ask | | Put Bid | | Put Ask | |
|---|---|---|---|---|---|---|---|---|
| | T Stat. | P Val. (%) | T Stat. | P Val. (%) | T Stat. | P Val. (%) | T Stat. | P Val. (%) |
| IVS-KPSVR | 0.39 | 69.62 | -0.43 | 66.74 | -0.63 | 52.87 | -0.71 | 47.81 |
| IVS-BKPSVR | 0.55 | 58.45 | -0.56 | 57.46 | -1.29 | 19.64 | -1.31 | 18.93 |
| IVS-EKPSVR | 0.43 | 66.67 | -0.82 | 41.23 | -1.08 | 28.15 | -1.09 | 27.43 |

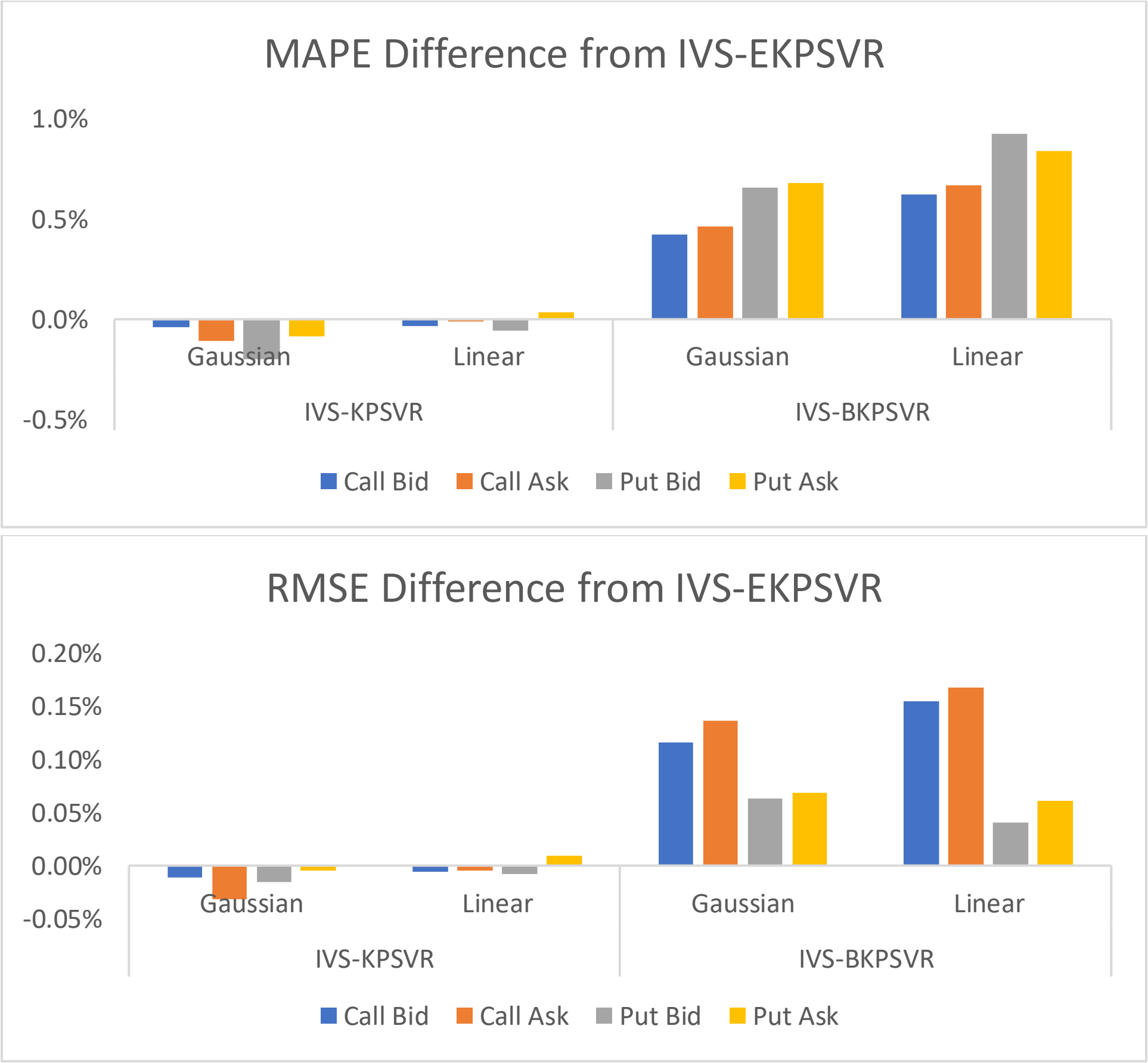

**Figure 8: Average Performance Difference from IVS-EKPSVR**

Edge points on the strike-maturity grid have less reference points to infer their values, in which case accuracy on the edges is negatively impacted. In all IVS models, the edge points that lie outside of the 20 strike prices in the center part of the grid can increase the regression error by about 30 ~ 60% (detailed numbers can be found in the Appendix). In Figure 9, we present the performance differences of IVS-KPSVR and IVS-BKPSVR from IVS-EKPSVR if edge strike prices are ruled out. In the center part of the grid, IVS-EKPSVR again outperforms IVS-BKPSVR but shows identical results to IVS-KPSVR with much less support vectors.

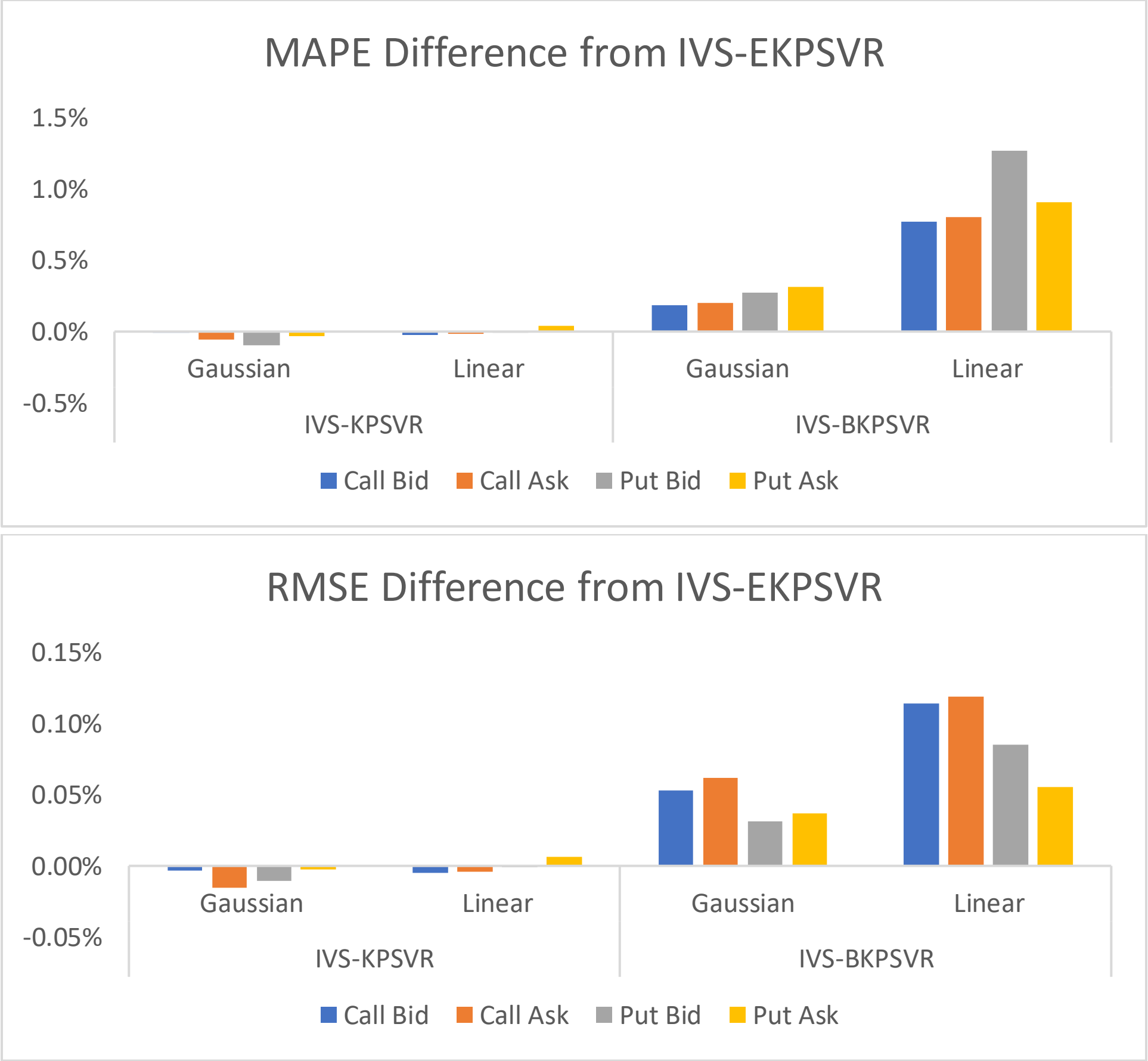

**Figure 9: Average Performance Difference from IVS-EKPSVR (without Edge Prices)**

Before Pegasos, two stochastic gradient descent based methods were introduced to solve SVM classification problems – Kivinen et al. (2004) and Zhang (2004). Kivinen suggests a learning rate of $\frac{p}{\lambda\sqrt{t}}$ in their algorithm called NORMA, while Zhang simply let it be a constant $\eta$, regardless of iterations (we name it BSGD). To adapt these methods in an online adaptive regression setting, we update Step 2.c in Algorithm 4 to $S[s] \leftarrow \left(1 - \frac{p}{\sqrt{t}}\right) S[s]$ (for NORMA) and $S[s] \leftarrow (1 - \eta\lambda) S[s]$ (for BSGD). Step sizes in Step 2.d are changed to $\pm \frac{p}{\lambda\sqrt{t}}$ and $\eta$ respectively. We name these enhanced algorithms IVS-NORMA[4] and IVS-BSGD. In IVS-BSGD, constant $\eta$ is set to 0.01 and $\lambda$ is set to 10 to attain a relatively good result that balances the support vector size and prediction error (tuned using data from 01/27/2014). Since the shrinkage multiplier for $S[s]$ are nonnegative in both cases, the warmup parameter is not needed for these two algorithms. All other parameters remain the same as in IVS-EKPSVR. Table 3 shows that, under Gaussian kernel, IVS-EKPSVR uses the least number of support vectors but also achieves the smallest error rates. This further verifies the superiority of the learning rate $\frac{1}{\lambda t}$ and all other enhancements behind IVS-EKPSVR.

---

[4] Optimal $p$ is determined by $0.5\sqrt{\left(2 + \frac{0.5}{\sqrt{T}}\right)}$, where $T$ is the maximum number of iterations. We let $T$ go to infinity due to the large size of option tick data and hence $p = 0.71$.

**Table 3: Performance Summary of Competing Algorithms**

| | Call Bid | | | Call Ask | | | Put Bid | | | Put Ask | | |
|---|---|---|---|---|---|---|---|---|---|---|---|---|
| | MAPE (%) | RMSE (%) | SV | MAPE (%) | RMSE (%) | SV | MAPE (%) | RMSE (%) | SV | MAPE (%) | RMSE (%) | SV |
| IVS-EKPSVR | **12.09** | **2.27** | 110 | **11.86** | **2.48** | **104** | **14.58** | **1.63** | **116** | **12.45** | **1.66** | **120** |
| IVS-NORMA | 18.52 | 3.29 | 170 | 17.89 | 3.34 | 169 | 24.66 | 2.72 | 164 | 21.13 | 2.63 | 160 |
| IVS-BSGD | 18.63 | 3.38 | **105** | 17.20 | 3.39 | 106 | 27.35 | 3.01 | 120 | 23.10 | 2.87 | 125 |

Hyper parameter tuning directly impacts the number of support vectors and prediction accuracy. Parameter $\gamma$ in the Gaussian kernel controls how far the influence of a support vector reaches; constant $\rho$ is related to local fitness; parameter $\lambda$ regularizes the loss function; warmup coefficient $\omega$ defines the magnitude of model updates at each step. In the following analysis, we discuss the sensitivity of IVS-EKPSVR to these parameters using a Gaussian kernel and data from 01/27/2014. All the parameters chosen previously are based on the subsequent grid search process that trades off the model complexity (the number of support vectors) and error rates.

Figure 10 demonstrates the changes of average support vector size and MAPE of the four models (Call Bid, Call Ask, Put Bid and Put Ask) with regard to $1/\gamma$ and $\rho$. A larger $1/\gamma$ exerts a greater influence of support vectors to more distant samples, while a smaller value constrains support vector's influence on nearby data points and hence calls for more support vectors in a model. The MAPE, on the other hand, reaches its lowest value when $1/\gamma = 4$. Local fitness threshold $\rho$ mainly controls the number of support vectors. The larger the threshold $\rho$ is, the more support vectors will be selected. In the MAPE plot, given a fixed $1/\gamma$, parameter $\rho$ yields fairly stable error rates across different $\rho$ values. The spike in the MAPE plot occurs as a resultant of a large $1/\gamma$ and a small $\rho$, which produces very few support vectors in the model (a sudden drop at the bottom in the left figure), followed by a large MAPE.

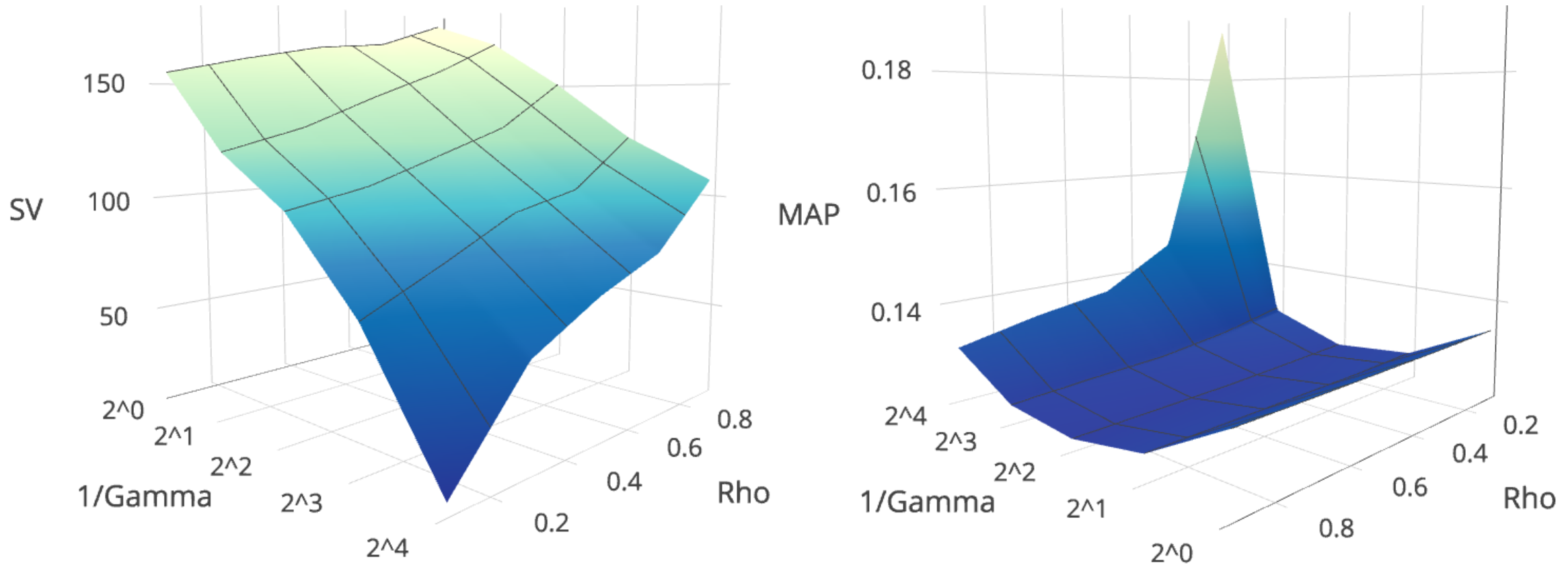

**Figure 10[5]: Sensitivity of Support Vector Size and MAP to Gamma ($\gamma$) and Rho ($\rho$)**

Figure 11 presents the sensitivity analysis of $\omega$ and $\lambda$. The warmup factor $\omega$ does not influence the MAPE to a great extent, but impacts the support vector sizes: a larger $\omega$ leads to a smaller support vector set. A smaller regularization parameter $\lambda$ puts more emphasis on the regression error by introducing more freedom or support vectors into the model, which further yields a reduced MAPE.

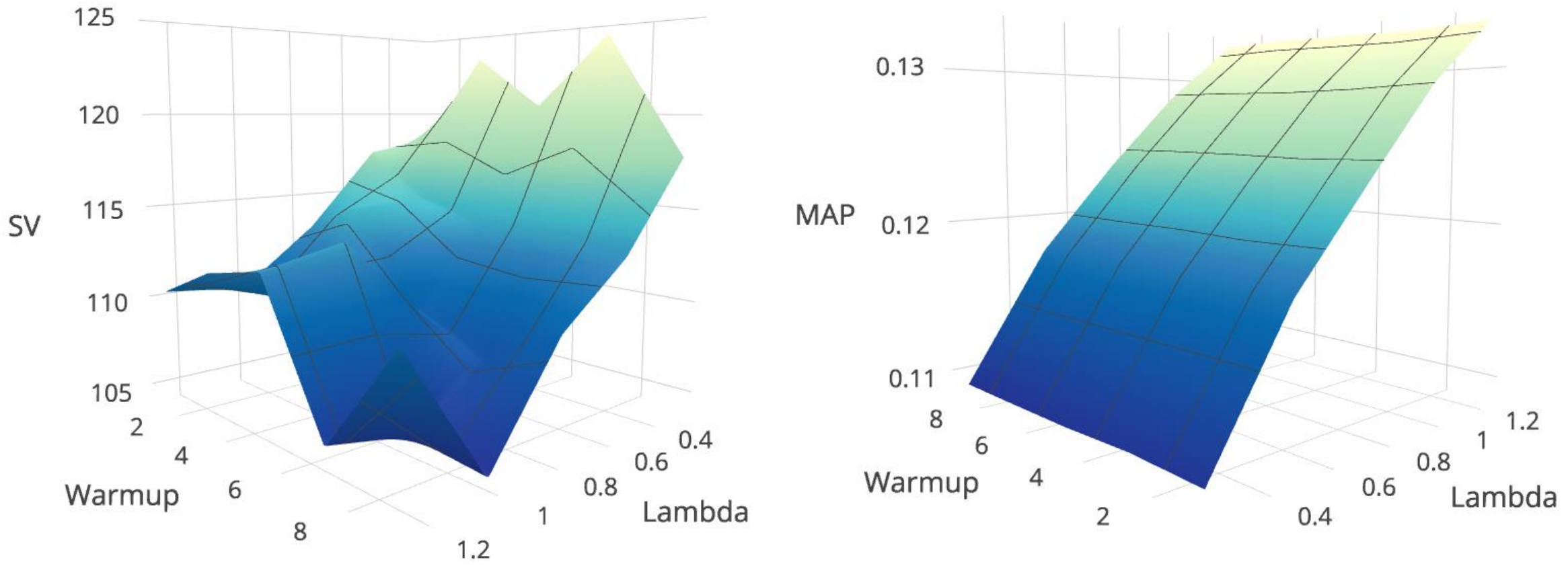

**Figure 11[5]: Sensitivity of Support Vector Size and MAP to Warmup ($\omega$) and Lambda ($\lambda$)**

In Figure 12, we contrast the runtime performance of FPGA against CPU implementations for the four parts of the algorithm as explained in Section 4.5. Given the support vector size limit of 200 (the original grid size), FPGA cannot fully utilize its computational power and does not show much speed improvement due to the communication overhead between CPU and FPGA. For this reason, we enlarge the support vector size 100 times to 20,000, which represents a much finer grid and a reasonable industry scale. In the following analysis, it is assumed that all data points on the grid are used as support vectors. During the model training phase, FPGA shows a 16.7 speedup for calculating a local fitness, 7.2 faster for completing the matrix inverse calculations upon a support vector addition and 5.4 speedup for a support vector removal. Prediction phase (Step 2.e in Algorithm 4) enjoys the most speedup due to its highly parallel nature. It obtains a 131.8 acceleration for predicting 20,000 samples. For a large-scale online implementation of machine learning algorithms, not only ours, FPGA technology is thus an excellent alternative to reduce latency and to enable agile responses to the unremitting changes in the real world.

[5] For a better viewing angle, the axis directions are different in the two plots.

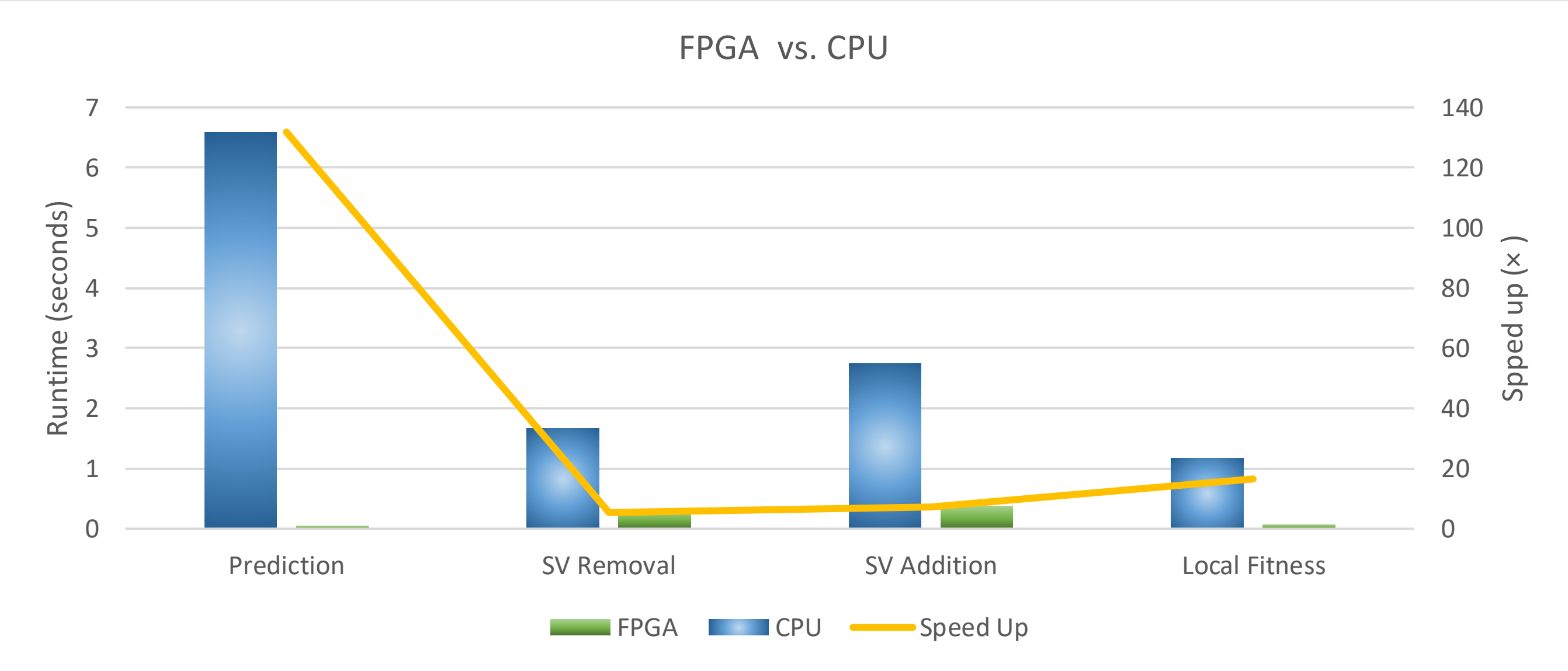

**Figure 12: FPGA vs. CPU speed comparison**

### 5.3 Online vs. offline comparison

In this section we compare our online algorithm with classic methods linear regression, gradient boosted trees, random forest and SVR used in the offline setting. We expect these offline methods to perform better given looser time constraints. In the following, our results are compared with the four offline methods in terms of performance and average runtime (Table 4). In particular, we use statsmodels module in Python for linear regression and scikit-learn module for random forest, gradient boosted trees and offline SVR. The runtime column measures the average time for all four Call/Put Bid/Ask models per update (including both training and inference). All offline algorithms apply the default configuration in the underlying module except that 20 trees are chosen for random forest (based on cross-validation using data from 01/27/2014) and offline SVR employs the same parameter values (where applicable) as our online SVR. All results are obtained using out-of-sample data from 01/28/2014 to 01/31/2014. The best numbers are indicated in bold in the table. We observe that the tree-based models generally have the best MAPE and RMSE but are also much more computationally intensive. The most important finding is that when it comes to performance measures the online version is inferior but it is very competitive (the gaps are not substantial). On the other hand, the online version is substantially faster. We can definitely conclude that the online algorithm is a much better choice when it comes to the trade-off between the two.

**Table 4: Performance Summary of Competing Offline Algorithms**

| | Call Bid | | Call Ask | | Put Bid | | Put Ask | | Runtime (ms) |
|---|---|---|---|---|---|---|---|---|---|
| | MAPE (%) | RMSE (%) | MAPE (%) | RMSE (%) | MAPE (%) | RMSE (%) | MAPE (%) | RMSE (%) | |
| IVS-EKPSVR | 12.09 | 2.27 | 11.86 | 2.48 | 14.58 | **1.63** | 12.45 | **1.66** | **7.58** |
| Linear Regression | 14.4 | 1.92 | 14.04 | 1.96 | 16.53 | 1.83 | 14.77 | 1.89 | 25.92 |
| Random Forest | 8.70 | **1.44** | 8.37 | 1.44 | **12.22** | 1.65 | **11.24** | 1.72 | 93.02 |
| Boosted Trees | **8.43** | **1.44** | **8.10** | **1.43** | 12.28 | 1.66 | 11.27 | 1.74 | 71.48 |
| Offline SVR | 25.47 | 3.50 | 23.70 | 3.39 | 37.95 | 4.13 | 35.12 | 4.13 | 22.21 |

## 6. Conclusion

This paper presents the first implementation of an online adaptive primal SVR algorithm with an application to model the implied volatility surface in the E-mini S&P 500 options market. We introduce feature vector selection and budget maintenance to control the number of support vectors and dynamically update the model once a new pattern or changed pattern emerges, which then evolves into the IVS-EKPSVR algorithm that outperforms either IVS-KPSVR or IVS-BKPSVR. We find that the linear kernel does not work well with FVS in regulating the support vector size, but performs similarly to Gaussian kernel in terms of error rates if an identical number of support vectors are used such as in IVS-KPSVR and IVS-BKPSVR. Due to less reference points, edges of the maturity-strike grid possess larger regression errors that may boost overall MAPE and RMSE by around 30 ~ 60%. Compared with competing methods, IVS-EKPSVR outperforms IVS-NORMA and IVS-BSGD to a great extent, with a MAPE gap up to 12.7% and RMSE gap up to 1.4%. In addition, FPGA hardware has been proved to significantly accelerate the training and prediction phase of our algorithm, up to 132x. Finally, IVS-EKPSVR reveals a good speed-performance balance in comparison with classic offline machine learning algorithms. Future work can focus on improving the prediction accuracy on the edges of the IVS grid, for example, by introducing more predictor variables from the markets.

## Acknowledgement

This work is supported by CME Group. We thank them for providing us with valuable data and we are grateful for their donation of the Maxeler FPGA hardware. We especially acknowledge Ryan Eavy, Executive Director, Architectures at CME Group for his support and introduction to Maxeler.

## Appendix

Table 5 shows that Feb 2014 maturity presents the most volatile properties with the largest standard deviation (std.), skewness (skew.) and kurtosis (kurt.). The longer the maturity is, the smaller the std. and the absolute values of skew. and kurt. are. Table 6 summarizes the performances of our algorithms while Table 7 shows the same except without edge strike prices.

**Table 5: Summary statistics of implied volatility on 01/27/2014**

| | Maturity | Mean (%) | Std. (%) | Skew. | Kurt. |
|---|---|---|---|---|---|
| Call Bid | Feb 2014 | 11.74 | 7.98 | -2.51 | 6.11 |
| | Mar 2014 | 14.42 | 1.86 | -2.37 | 14.73 |
| | April 2014 | 13.66 | 0.98 | -0.06 | -0.56 |
| | May 2014 | 14.47 | 0.81 | -0.03 | -0.64 |
| | June 2014 | 15.05 | 0.78 | -0.07 | -0.43 |
| Call Ask | Feb 2014 | 14.10 | 5.20 | -3.57 | 15.40 |
| | Mar 2014 | 15.03 | 1.48 | -0.86 | 4.72 |
| | April 2014 | 14.02 | 0.96 | -0.03 | -0.55 |
| | May 2014 | 14.80 | 0.79 | 0.02 | -0.59 |
| | June 2014 | 15.34 | 0.77 | -0.04 | -0.42 |
| Put Bid | Feb 2014 | 14.53 | 2.30 | 0.28 | -0.95 |
| | Mar 2014 | 14.40 | 1.54 | -0.21 | -0.67 |
| | April 2014 | 15.06 | 1.02 | -0.02 | -0.66 |
| | May 2014 | 14.88 | 0.80 | -0.01 | -0.67 |
| | June 2014 | 14.97 | 0.82 | 0.03 | -0.50 |
| Put Ask | Feb 2014 | 15.30 | 2.35 | 0.10 | -1.01 |
| | Mar 2014 | 14.83 | 1.53 | -0.25 | -0.58 |
| | April 2014 | 15.38 | 1.03 | -0.01 | -0.72 |
| | May 2014 | 15.19 | 0.82 | 0.00 | -0.74 |
| | June 2014 | 15.24 | 0.84 | 0.04 | -0.57 |

**Table 6: Performance summary of our algorithms (in %)**

| | | Call Bid | | Call Ask | | Put Bid | | Put Ask | |
|---|---|---|---|---|---|---|---|---|---|
| | Kernel | MAPE | RMSE | MAPE | RMSE | MAPE | RMSE | MAPE | RMSE |
| IVS-KPSVR | Gaussian | 12.05 | 2.26 | 11.76 | 2.45 | 14.38 | 1.61 | 12.36 | 1.66 |
| | Linear | 12.20 | 2.08 | 11.70 | 2.14 | 14.42 | 1.63 | 12.37 | 1.67 |
| IVS-BKPSVR | Gaussian | 12.51 | 2.39 | 12.33 | 2.61 | 15.24 | 1.69 | 13.12 | 1.73 |
| | Linear | 12.86 | 2.24 | 12.38 | 2.31 | 15.40 | 1.68 | 13.18 | 1.72 |
| IVS-EKPSVR | Gaussian | 12.09 | 2.27 | 11.86 | 2.48 | 14.58 | 1.63 | 12.45 | 1.66 |
| | Linear | 12.23 | 2.08 | 11.71 | 2.14 | 14.48 | 1.63 | 12.34 | 1.66 |

**Table 7: Performance summary of our algorithms without edge strikes (in %)**

| | | Call Bid | | Call Ask | | Put Bid | | Put Ask | |
|---|---|---|---|---|---|---|---|---|---|
| | Kernel | MAPE | RMSE | MAPE | RMSE | MAPE | RMSE | MAPE | RMSE |
| IVS-KPSVR | Gaussian | 8.13 | 1.47 | 7.80 | 1.50 | 9.90 | 1.23 | 8.41 | 1.21 |
| | Linear | 8.50 | 1.50 | 8.13 | 1.51 | 10.16 | 1.27 | 8.68 | 1.25 |
| IVS-BKPSVR | Gaussian | 8.32 | 1.53 | 8.06 | 1.58 | 10.27 | 1.27 | 8.76 | 1.25 |
| | Linear | 9.30 | 1.62 | 8.95 | 1.63 | 11.44 | 1.36 | 9.55 | 1.30 |
| IVS-EKPSVR | Gaussian | 8.14 | 1.48 | 7.85 | 1.51 | 9.99 | 1.24 | 8.45 | 1.21 |
| | Linear | 8.53 | 1.51 | 8.15 | 1.51 | 10.17 | 1.28 | 8.64 | 1.25 |